\documentclass[10pt,conference]{IEEEtran}
\usepackage{cite}
\usepackage{amsmath,amssymb,amsfonts}
\usepackage{algorithmic}
\usepackage{graphicx}
\usepackage{textcomp}
\usepackage[dvipsnames]{xcolor}
\usepackage{soul}
\usepackage[hyphens]{url}
\usepackage{fancyhdr}
\usepackage[bookmarks=true,breaklinks=true,letterpaper=true,colorlinks,citecolor=blue,linkcolor=blue,urlcolor=blue]{hyperref}
\usepackage{tcolorbox}
\usepackage{enumitem}

\newcommand{\hpcayear}{2025}

\usepackage{tikz}
\newcommand*\circled[1]{\tikz[baseline=(char.base)]{\node[shape=circle,fill,inner sep=0.5pt] (char) {\textcolor{white}{#1}};}}

\usepackage{mathtools}
\DeclarePairedDelimiter{\ceil}{\lceil}{\rceil}

\usepackage{listings}
\definecolor{codegray}{rgb}{0.9, 0.9, 0.9}
\definecolor{codegreen}{rgb}{0, 0.6, 0}
\definecolor{codeblack}{rgb}{0, 0, 0}
\lstdefinestyle{mystyle}{
    language=bash,
    backgroundcolor=\color{codegray},
    commentstyle=\color{codegreen},
    frame=single, 
    framerule=0.75pt,
    rulecolor=\color{codeblack},
    columns=fullflexible,
    basicstyle={\linespread{1}\rmfamily\small},
    lineskip=0.5ex, 
    xleftmargin=.15cm, xrightmargin=.12cm, 
    aboveskip=.15cm, belowskip=.24cm, 
    framextopmargin=.075cm, framexbottommargin=.05cm,
    breaklines, breakautoindent=false, breakindent=2ex, 
}
\usepackage[ruled, linesnumbered, vlined]{algorithm2e}
\usepackage{setspace}
\SetAlFnt{\footnotesize}
\SetAlCapNameFnt{\small}
\SetAlCapFnt{\small}
\IncMargin{3pt}
\SetAlgoNlRelativeSize{-0.5}
\SetKwProg{Def}{def}{$\,$:}{}
\SetKwProg{Func}{Func}{$\,$:}{}
\SetKwInOut{Input}{Input}
\SetKwInOut{Output}{Output}

\usepackage{etoolbox} 
\makeatletter
\patchcmd{\algocf@makecaption@ruled}{\hsize}{\textwidth}{}{} 
\patchcmd{\@algocf@start}{-1.5em}{0em}{}{} 
\makeatother

\usepackage{booktabs, makecell, multirow, tabularx, threeparttable} 
\usepackage{caption}
\definecolor{yuz_colour}{RGB}{191, 232, 255}

\definecolor{george_colour}{RGB}{240, 180, 100}

\definecolor{mar_colour}{RGB}{216, 191, 216}

\newcommand{\hpcasubmissionnumber}{NaN}
\title{
BitMoD: Bit-serial Mixture-of-Datatype \\LLM Acceleration
}

\def\hpcacameraready{} 

\newcommand\hpcaauthors{
    Yuzong Chen$^\dagger$, Ahmed F. AbouElhamayed$^\dagger$, Xilai Dai$^\dagger$, Yang Wang$^\ddagger$, Marta Andronic$^\mathsection$, \\George A. Constantinides$^\mathsection$, and Mohamed S. Abdelfattah$^\dagger$
}
\newcommand\hpcaaffiliation{
    $^\dagger$Computer Systems Lab, Cornell University \\$^\ddagger$Systems and Networking Research Group, Microsoft Research
    \\$^\mathsection$Department of Electrical and Electronic Engineering, Imperial College London
}
\newcommand\hpcaemail{ 
    $^\dagger$\{yc2367, afa55, xd44, mohamed\}@cornell.edu
    \\$^\ddagger$yang.wang92@microsoft.com 
    \\$^\mathsection$\{marta.andronic18, g.constantinides\}@imperial.ac.uk
}

\author{
  \ifdefined\hpcacameraready
      \IEEEauthorblockN{\hpcaauthors{}\vspace{3pt}}
      \IEEEauthorblockA{
        \hpcaaffiliation{} \vspace{3pt} \\
        \hpcaemail{}
        \vspace{-5pt}
      }
  \else
      \IEEEauthorblockN{\normalsize{HPCA \hpcayear{} Submission
      \textbf{\#\hpcasubmissionnumber{}}} \\
      \IEEEauthorblockA{
        Confidential Draft. Do NOT Distribute!!
      }
    }
  \fi 
}

\fancypagestyle{camerareadyfirstpage}{%
  \fancyhead{}
  
  \fancyhead[C]{
    \ifdefined\aeopen
    \parbox[][12mm][t]{13.5cm}{\hpcayear{} IEEE International Symposium on High-Performance Computer Architecture (HPCA)}    
    \else
      \ifdefined\aereviewed
      \parbox[][12mm][t]{13.5cm}{\hpcayear{} IEEE International Symposium on High-Performance Computer Architecture (HPCA)}
      \else
      \ifdefined\aereproduced
      \parbox[][12mm][t]{13.5cm}{\hpcayear{} IEEE International Symposium on High-Performance Computer Architecture (HPCA)}
      \else
      \parbox[][0mm][t]{13.5cm}{\hpcayear{} IEEE International Symposium on High-Performance Computer Architecture (HPCA)}
    \fi 
    \fi 
    \fi 
    \ifdefined\aeopen 
      \includegraphics[width=12mm,height=12mm]{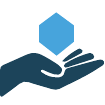}
    \fi 
    \ifdefined\aereviewed
      \includegraphics[width=12mm,height=12mm]{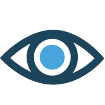}
    \fi 
    \ifdefined\aereproduced
      \includegraphics[width=12mm,height=12mm]{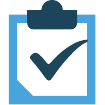}
    \fi
  }
  \fancyfoot[C]{}
}
\ifdefined\hpcacameraready 
\else
\fi

\ifdefined\eaopen

\fi

\begin{document}
\maketitle
\begin{abstract}

Large language models (LLMs) have demonstrated remarkable performance across various machine learning tasks. 
Yet the substantial memory footprint of LLMs significantly hinders their deployment.
In this paper, we improve the accessibility of LLMs through BitMoD\footnote{Code is available at: https://github.com/yc2367/BitMoD-HPCA-25}, an algorithm-hardware co-design solution that enables efficient LLM acceleration at low weight precision. 
On the algorithm side, BitMoD introduces fine-grained data type adaptation that uses a different numerical data type to quantize a group of (e.g., 128) weights.
Through the careful design of these new data types, BitMoD is able to quantize LLM weights to very low precision (e.g., 4 bits and 3 bits) while maintaining high accuracy. 
On the hardware side, BitMoD employs a bit-serial processing element to easily support multiple numerical precisions and data types; our hardware design includes two key innovations:
First, it employs a unified representation to process different weight data types, thus reducing the hardware cost. 
Second, it adopts a bit-serial dequantization unit to rescale the per-group partial sum with minimal hardware overhead. 
Our evaluation on six representative LLMs demonstrates that BitMoD significantly outperforms state-of-the-art LLM quantization and acceleration methods. 
For discriminative tasks, BitMoD can quantize LLM weights to 4-bit with $<\!0.5\%$ accuracy loss on average. 
For generative tasks, BitMoD is able to quantize LLM weights to 3-bit while achieving better perplexity than prior LLM quantization scheme. 
Combining the superior model performance with an efficient accelerator design, BitMoD achieves an average of $1.69\times$ and $1.48\times$ speedups compared to prior LLM accelerators ANT and OliVe, respectively.
\end{abstract}

\section{Introduction}

Large language models (LLMs) have achieved significant breakthroughs in natural language processing tasks~\cite{opt, llama}. However, the growth of LLM size and complexity continues to outpace the scaling of compute performance and memory capacity in existing hardware platforms~\cite{AI-mem-wall}. For example, the first generation of the GPT model, introduced in 2018, contains only 117 million parameters, while the second and third generations grew more than $10\times$ and $1000\times$, respectively within two years~\cite{GPT3}. This rapid increase in size necessitates significant memory capacity for model deployment, hindering their wide adoption, especially in edge scenarios with limited compute and memory resources. For instance, the state-of-the-art (SOTA) open-source LLM family, \text{Llama-3}~\cite{llama-3}, contains more than 8 billion parameters and requires more than 16GB of memory to store the model weights in 16-bit floating-point (\texttt{FP16}) format, which cannot fit in an edge GPU such as Jetson-TX2 with 8GB memory~\cite{jetson-gpu}. Therefore, designing novel LLM compression algorithms, together with accelerators co-designed for efficient deployment of the compressed models, presents a promising solution to enhancing the accessibility of LLMs on edge devices.

Quantization serves as one of the most hardware-efficient methods to mitigate the computation and memory demands of LLMs. Generally, there are two types of quantization mechanisms. The first one is quantization-aware training (QAT), where retraining is needed to update model weights and quantization parameters (e.g., scaling factors)~\cite{ant, llm-qat}. The second approach is post-training quantization (PTQ), which does not require retraining~\cite{smoothquant, gptq, awq, students, omniquant, olive, quip}. Although QAT can achieve more competitive accuracy than PTQ, the prohibitive cost of retraining LLMs makes it less practical. As a result, PTQ is commonly adopted in existing LLM quantization studies. While some PTQ works quantize both weights and activations into low precision~\cite{smoothquant, omniquant, olive}, weight-only quantization can offer a better trade-off between model accuracy and hardware efficiency for edge deployment of LLMs, where weights dominate the memory footprint~\cite{gptq, awq, students, quip}. 
However, existing weight-only quantization works on GPUs suffer from poor computational efficiency since GPUs lack dedicated hardware to perform multiplication between integer weight and floating-point activation. Consequently, these methods must first dequantize the weight to FP16 and rely on the floating-point pipeline for computation. 

To achieve better computational efficiency for LLMs, a recent accelerator work, FIGNA~\cite{figna}, proposes a family of dedicated computing units for mixed-precision arithmetic between integer weights and floating-point activations. 
To further unleash the potential of quantization for improved hardware efficiency, several works have proposed algorithm-hardware co-design solutions based on \textit{custom} low-precision data types~\cite{microscaling, mx-format, ant, olive}. 
The microscaling format (MX)~\cite{microscaling, mx-format}, assigns 8-bit metadata as the shared exponent to a group of low-precision weights. 
ANT~\cite{ant} introduces a new data type that better adapts to the intra-tensor value distribution, thus reducing the quantization error. 
OliVe~\cite{olive} proposes an outlier-victim-pair quantization mechanism, where an outlier value with a large magnitude is represented with an ``Adaptive Biased Float" format and can be protected by pruning its adjacent victim value that has a small magnitude. 


In this paper, we propose \textit{BitMoD}\footnote{\textit{BitMoD} stands for \underline{Bit}-serial computation with \underline{M}ixture \underline{o}f \underline{D}ata types.}, 
an algorithm-hardware co-design solution for efficient LLM acceleration at low weight precision. On the algorithm side, \textit{BitMoD} exploits the \textit{per-group} quantization~\cite{group-quant}, 
and modifies low-precision floating-point data types by repurposing the redundant zero value with a special value, which provides the ability to better adapt the data type itself to the numerical distribution of each weight group. Through careful choice of special values, \textit{BitMoD} is able to quantize LLM weights to very low precision (e.g., 4-bit and 3-bit) with tiny encoding overhead while maintaining good model accuracy. On the hardware side, \textit{BitMoD} employs the bit-serial computing paradigm with a unified representation for different low-precision data types to efficiently trade-off weight precision and hardware efficiency. 


The main contributions of this paper are summarized below:
\begin{enumerate}
    \item We propose \textit{BitMoD}, a hardware-efficient PTQ solution for LLM acceleration. \textit{BitMoD} introduces new data types that are tailored for per-group weight quantization at 4-bit and 3-bit precision with tiny encoding overhead. 
    \item We demonstrate that the proposed data types can be seamlessly integrated with other quantization optimization techniques, achieving better model perplexity than SOTA software-only LLM quantization works. 
    \item We propose an efficient accelerator design for \textit{BitMoD}, which adopts a unified bit-serial representation for multiple low-precision data types. This effectively reduces the hardware cost to perform computation between low-precision weights and FP16 activations, and trades-off weight precision for improved hardware efficiency.
    \item Our evaluation on six representative LLMs shows that on average, \textit{BitMoD} achieves $2.2\times$ speedup and  $2.31\times$ better energy efficiency compared to the baseline FP16 accelerator, \textit{without} loss in accuracy. Compared to SOTA accelerators ANT and OliVe, \textit{BitMoD} achieves an average speedup of $1.69\times$ and $1.48\times$, respectively. 
\end{enumerate}
%

\section{Background and Motivation}
\subsection{Why Weight Quantization for LLMs?}
\begin{figure}
        \centering
        \includegraphics[width=1\linewidth]{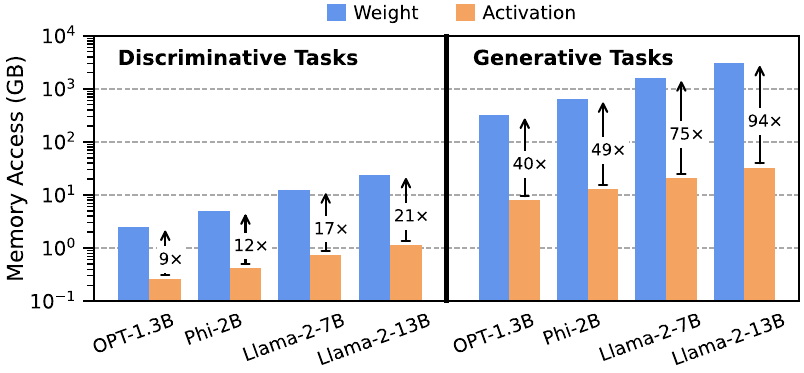}
        \vspace{-12pt}
        \caption{
        Total memory access of weights and activations on discriminative tasks (with 256 input tokens and 1 output token) and generative tasks (with 256 input tokens and 256 generated tokens). Note the log scale on the y-axis. Note that the gap between weight and activation memory accesses increases for generative tasks at batch size 1 despite a much larger KV-cache than discriminative tasks. While prior work~\cite{flexgen} has correctly reported a memory bottleneck caused by the KV-cache, this only occurs for 175B+ parameter models with a high batch size (e.g., 512) and a context lengths exceeding 512 tokens. This scenario is less relevant to our focus on low-batch edge LLM inference where the weights indeed dominate the total memory accesses.
        }
        \label{fig:mem_footprint}
        \vspace{-3pt}
    \end{figure}

To demonstrate the importance of LLM weight quantization for edge applications, we profile the total memory access footprint of weight and activation for four representative LLMs running both discriminative and generative tasks with a batch size of 1. For discriminative tasks, the LLM receives an input context and outputs a single token such as in sentiment analysis~\cite{sst2} and multiple-choice question answering~\cite{arc-c}. For generative tasks, the LLM receive an input context and output multiple tokens. We set the input to output sequence length to 256\,:\,1 and 256\,:\,256 for discriminative and generative tasks, respectively, catering for edge applications as suggested by Lin \textit{et al.}~\cite{awq}.
As shown in Fig. \ref{fig:mem_footprint}, the LLM weights access consumes orders of magnitude larger memory than the activations access. Although discriminative tasks only need to output a single token (e.g., ``A''/``B''/``C'' for multiple-choice question answering), the weight tensor dimension of an LLM (e.g., 2048 for OPT-1.3B) is much larger than the input token length, leading to memory access dominated by weights. Moreover, generative tasks necessitate repeated weight fetching for every new output token, resulting in significantly higher memory access for LLM weights. Thus, weight quantization is more effective for deploying LLMs in edge scenario where the batch size is small and the input token length is typically short. 

\subsection{Quantization Basics}
One of the most popular quantization schemes is integer quantization, where a floating-point value is scaled and rounded to a low-precision integer. There are two widely used quantization modes -- \textit{symmetric} and \textit{asymmetric}. Symmetric integer quantization can be expressed as follows:
    \begin{equation}
        \Delta = \frac{ W_{f{\text{max}}} }{ 2^{b-1} - 1 }  ; 
        \  W_q = \texttt{Round}\left(\frac{ W_f }{ \Delta }\right) ; 
        \  W_{qf} = W_q \cdot \Delta
        \label{eq:quant_sym}
    \end{equation} 
where $W_{f}$ is the original floating-point tensor, $W_{f{\text{max}}}$ is the absolute maximum value, $b$ is the quantized integer precision, $\Delta$ is the scaling factor, $W_q$ is the quantized integer value, and $W_{qf}$ is the floating-point value after performing dequantization (i.e., re-scaling).




The symmetric quantization assumes that the minimum and maximum values of a tensor have the same absolute value (i.e., symmetric value range), but this is not always true. Hence, another popular mode of quantization is asymmetric quantization, which can be expressed as follows:
    \begin{equation*}
        \Delta = \frac{ \texttt{Range}\left(W_f\right) }{ 2^b - 1 } \, ; 
        \ \  z = \texttt{Round}\left(\frac{ -W_{f{\text{min}}} }{ \Delta }\right)
    \end{equation*} 
    \begin{equation}
        W_q = \texttt{Round}\left(\frac{W_f}{\Delta}\right) + z \, ; 
        \ \  W_{qf} = \left(W_q - z\right) \cdot \Delta 
        \label{eq:quant_asym}
    \end{equation} 
where $W_{f{\text{min}}}$ is the absolute minimum value of $W_{f}$, and $z$ represents the zero-point of the quantized tensor. 



\subsection{Motivation} \label{sec:motivation}
We analyze several techniques that are widely adopted in recent quantization studies, which motivates our proposed \textit{BitMoD} framework. We mainly focus on weight quantization in our discussion. 


\vspace{4pt}
\noindent
\textbf{Quantization Granularity Matters. } 
Consider a floating-point weight tensor $W_f^{K\times D}$, where $K$ represents the number of output channels and $D$ is the channel size. There are three \textit{granularities} to quantize the model weight: per-tensor, per-channel, and per-group.
The \textit{per-tensor} quantization uses the same scaling factor to quantize a whole weight tensor, while \textit{per-channel} quantization divides the weight tensor along the output channel into $K$ vectors, and quantizes every vector $W_f^{1\times D}$ independently. 
However, given the large tensor size and hidden dimension of LLMs, these two granularities still lead to large quantization error. Specifically, the quantization error of a dequantized weight in Eq. \ref{eq:quant_sym} can be expressed as:
    \begin{equation}
        \texttt{Error}(W_{qf}) = \texttt{ErrorRound}\left(\frac{ W_f }{ \Delta }\right) \cdot \Delta
    \end{equation} 
where \texttt{ErrorRound} is the rounding error during quantization, which has been shown to have an expected value of 0.25~\cite{awq}. 
Therefore, the quantization error is proportional to the scaling factor $\Delta$, which is further proportional to the maximum value and range for symmetric (Eq.~\ref{eq:quant_sym}) and asymmetric (Eq.~\ref{eq:quant_asym}) quantization, respectively.

    \begin{figure} [t]
        \centering
        \includegraphics[width=1\linewidth]{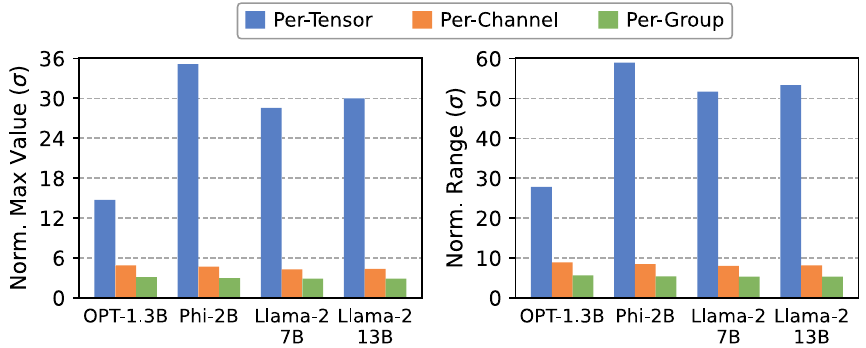}
        \vspace{-12pt}
        \caption{Maximum value and value range for different quantization granularity. Results are normalized to the standard deviation ($\sigma$) of the weight vector at the corresponding granularity, then averaged across all weight vectors. The per-group granularity has a group size of 128. }
        \label{fig:quant_granularity}
    \end{figure}
    
In order to further reduce the quantization error, recent LLM quantization studies adopt the \textit{per-group} granularity~\cite{group-quant, gptq, awq, students}. The per-group quantization further divides a weight channel $W_f^{1\times D}$ into $D/G$ groups, each with a group size of $G$. The group size introduces extra overhead to store the quantization parameters, i.e., scaling factor (and zero-point) for every group, and is usually set to 128 in SOTA quantization frameworks to balance accuracy and memory overhead~\cite{gptq, awq}. Fig.~\ref{fig:quant_granularity} demonstrates the benefits of per-group quantization by showing the maximum value and range in four representative LLMs at different granularities. The per-group granularity has the lowest maximum value and range, hence will have a lower quantization error compared to the other two granularity. Therefore, we focus on per-group quantization in this work.

\vspace{4pt}
\noindent
\textbf{Quantization Data Type Matters. } 
Numerous studies have proposed custom data types for quantization at the per-channel granularity~\cite{ant, olive, microscaling}. We analyze the effects of adopting different data types for per-channel and per-group weight quantization. We explore four basic data types at 4-bit precision: integer with symmetric (\texttt{INT4-Sym}) and asymmetric (\texttt{INT4-Asym}) quantization, floating-point (\texttt{FP4}), and the \texttt{Flint} data type proposed by ANT~\cite{ant}. Table~\ref{tab:dtype_4b_comp} shows the resulting perplexity on the Wikitext-2 dataset \cite{wikitext}. We highlight two important observations. 
First, although \texttt{Flint} can achieve better perplexity at the per-channel granularity, it never outperforms other data types at the per-group granularity. 
Second, the per-group \texttt{INT4-Asym} and \texttt{FP4} quantization achieve the best perplexity on some but not all studied LLMs, indicating that both asymmetry and \texttt{FP} data types are favorable for per-group quantization. 
The reason behind this is twofold. First, weight tensors typically exhibit Gaussian-like distribution that fits well to the floating-point data type~\cite{fp6-llm, nf4}. Second, while the effects of outliers are mitigated by per-group quantization, a weight group can still contain outliers in an asymmetric pattern, being either solely positive or negative, as highlighted in previous studies~\cite{llm-int8, group-quant}. This characteristic benefits from asymmetric quantization.

    \begin{table} [t]
      \centering
      \setlength{\tabcolsep}{4pt}
      \renewcommand{\arraystretch}{1.2}
      \footnotesize
      \caption{Wikitext-2 perplexity ($\downarrow$) under different quantization granularity and 4-bit data types. ``PC'' and ``PG'' stand for per-channel and per-group, respectively. The group size is 128.}
      \vspace{-3pt}
        \begin{tabular}{ c | c  c | c  c | c  c | c  c }
            \Xhline{0.3ex}
              Model & \multicolumn{2}{c|}{OPT-1.3B} & \multicolumn{2}{c|}{Phi-2B} & \multicolumn{2}{c|}{Llama-2-7B}  & \multicolumn{2}{c}{Llama-2-13B} \\
            \cline{1-9}
              Granularity & PC & PG & PC & PG & PC & PG & PC & PG \\
            \Xhline{0.3ex}
              FP16 & 14.62 & 14.62 & 9.71 & 9.71 & 5.47 & 5.47 & 4.88 & 4.88 \\
            \hline
              INT4-Sym & 36.05 & 16.04 & 13.03 & 11.15 & 12.92 & 5.84 & 5.47 & 5.07 
            \\
              INT4-Asym & 48.41 & 15.41 & 12.08 & \textbf{10.67} & 8.89 & \textbf{5.77} & 5.27 & \textbf{5.01} 
            \\
              FP4 & 16.07 & \textbf{14.99} & \textbf{11.24} & 10.68 & 8.07 & \textbf{5.77} & \textbf{5.15} & 5.05 
            \\
              Flint & \textbf{15.87} & 16.23 & 11.71 & 11.23 & \textbf{6.67} & 6.09 & 5.31 & 5.29 \\
            \Xhline{0.3ex}
        \end{tabular}
        \vspace{-10pt}
      \label{tab:dtype_4b_comp}
    \end{table}
    
The above observation motivates us to explore new quantization data types that can combine the benefits of \textit{asymmetry} and \texttt{FP} formats to achieve better accuracy under per-group quantization. We notice that the basic \texttt{FP} data types have symmetric quantization values due to the inherent sign-magnitude binary representation that contains positive and negative zero values. Our key insight is that we can introduce additional asymmetry to \texttt{FP} by repurposing a redundant zero value with another \textit{special value}. This approach provides us with two key benefits. First, it allows to fully utilize the limited quantization levels. Although the redundant zero value does not affect high-precision formats such as \texttt{FP16}, it constitutes a large fraction of quantization levels at low precision (e.g., $12.5\%$ at 3-bit precision). Second, we can tune the special value to make the extended \texttt{FP} data types better adapt to the per-group weight distribution, which we discuss in Section~\ref{sec:dtype_adapt}.


\vspace{4pt}
\noindent
\textbf{Quantization Bit-width Matters. } 
While prior LLM accelerators mainly rely on bit-parallel architectures that support 8-bit and 4-bit precision~\cite{ant, olive, figna}, recent studies have shown that 6-bit floating-point weights exhibit negligible accuracy loss across various LLM models and tasks~\cite{zeroq_llm, fp6-llm}. Motivated by this, we analyze the effects of using different 6-bit data types for per-group LLM weight quantization. We consider four data types: integer with symmetric (\texttt{INT6-Sym}) and asymmetric (\texttt{INT6-Asym}) quantization, floating-point with 2-bit exponent and 3-bit mantissa (\texttt{FP6-E2M3}), and floating-point with 3-bit exponent and 2-bit mantissa (\texttt{FP6-E3M2}). Table~\ref{tab:dtype_6b_comp} compares the resulting perplexity of different quantization data types on Wikitext-2~\cite{wikitext} and C4~\cite{c4} datasets. On average, the studied 6-bit data types achieve similar and negligible perplexity loss compared to the \texttt{FP16} baseline. For example, the average perplexity loss of \texttt{INT6-Sym} is less than 0.05, and its simple integer representation offers a promising solution to efficient LLM acceleration. Therefore, it is crucial for an accelerator to support diverse quantization bit-width to offer a better trade-off between memory footprint and model accuracy. 

A natural solution for accommodating variable precision is to adopt bit-serial architectures~\cite{stripes, pragmatic, bitwave, bbs}. However, existing bit-serial accelerators mainly target the integer data type, which causes significant accuracy loss at 3-bit precision as we will show in Section~\ref{sec:accuracy}. Furthermore, these accelerators cannot leverage per-group quantization for improved accuracy. This is because per-group quantization assigns different scaling factors for different groups, necessitating a floating-point unit with large area overhead to dynamically dequantize the partial sum after computing the dot-product for every group. Thus, an efficient dequantization mechanism with low hardware cost is desirable. 

    \begin{table} [t]
      \centering
      \setlength{\tabcolsep}{4pt}
      \renewcommand{\arraystretch}{1.2}
      \footnotesize
      \caption{Wikitext-2 and C4 perplexity ($\downarrow$) under different 6-bit data types. We use per-group weight quantization with a group size of 128. 
      }
      \vspace{-3pt}
        \begin{tabular}{ c | c  c | c  c | c  c | c  c }
            \Xhline{0.3ex}
              Model & \multicolumn{2}{c|}{OPT-1.3B} & \multicolumn{2}{c|}{Phi-2B} & \multicolumn{2}{c|}{Llama-2-7B}  & \multicolumn{2}{c}{Llama-2-13B} \\
            \cline{1-9}
              Dataset & Wiki & C4 & Wiki & C4 & Wiki & C4 & Wiki & C4 \\
            \Xhline{0.3ex}
              FP16 & 14.62 & 14.72 & 9.71 & 12.74 & 5.47 & 6.97 & 4.88 & 6.47 \\
            \hline
              INT6-Sym & \textbf{14.51} & 14.80 & 9.85 & 12.82 & \textbf{5.49} & \textbf{6.99} & \textbf{4.89} & \textbf{6.46} 
            \\
              INT6-Asym & 14.61 & 14.78 & \textbf{9.76} & \textbf{12.8} & \textbf{5.49} & \textbf{6.99} & \textbf{4.89} & \textbf{6.46} 
            \\
              FP6-E2M3 & 14.59 & \textbf{14.76} & 9.85 & \textbf{12.8} & 5.52 & \textbf{6.99} & 4.92 & 6.49 
            \\
              FP6-E3M2 & 14.81 & 14.81 & 9.81 & 12.87 & \textbf{5.49} & 7.02 & \textbf{4.89} & 6.50 \\
            \Xhline{0.3ex}
        \end{tabular}
        \vspace{-10pt}
      \label{tab:dtype_6b_comp}
    \end{table}


\vspace{4pt}
\noindent
\textbf{Algorithm-Hardware Co-Design Matters. } 
Numerous frameworks have been proposed to accelerate LLM execution, as depicted in Table~\ref{tab:overall_comp}.
SOTA algorithmic solutions such as AWQ~\cite{awq} quantize LLM weights to low-precision integer while preserving high accuracy. Nevertheless, AWQ is optimized for LLM acceleration on GPUs, which lack dedicated mixed-precision computing unit. As a result, it converts the low-precision weights to FP16 and relies on the GPU floating-point pipeline for computation, resulting in poor computational efficiency. 

In contrast, ANT~\cite{ant}, OliVe~\cite{olive}, and FIGNA~\cite{figna} propose efficient bit-parallel accelerators for quantized model acceleration. But their precision is limited to 8-bit and 4-bit, which restricts the ability to utilize other precision (e.g., 6-bit) for a better accuracy-efficiency trade-off. Moreover, their accelerators do not natively support per-group quantization, which requires a floating-point unit to dynamically dequantize the per-group partial sum on the fly. While Microscaling~\cite{microscaling} accommodates diverse precision, it necessities a floating-point pipeline to handle the shared micro-exponent of a weight group, leading to higher energy consumption compared to other low-precision compute units. 
Furthermore, given the significant memory footprint of LLMs, it is desirable to explore sub-4-bit quantization while maintaining good model accuracy, which ANT, OliVe, and Microscaling do not address. As we will show in Section~\ref{sec:accuracy}, the custom quantization data types proposed by ANT, OliVe, and Microscaling fail to achieve better accuracy than the simple asymmetric integer quantization at 4-bit weight precision, and cause unacceptable accuracy loss at 3-bit weight precision under per-group quantization. The above limitation motivates us to propose an efficient LLM acceleration framework that supports a wide range of hardware-friendly bit-widths, while maintaining good accuracy at low weight precision.

\section{BitMoD Quantization Framework}

In this section, we present the \textit{BitMoD} quantization framework, which includes new data type families tailored for per-group quantization at 3-bit and 4-bit precision. Section~\ref{sec:asym_fp_type} describes our proposed data types that extend the basic floating-point data types at 3-bit and 4-bit precision. Section~\ref{sec:dtype_adapt} presents an enhanced per-group LLM quantization strategy using the proposed data types. Section~\ref{sec:dequant} describes the hardware-efficient per-group dequantization mechanism using integer scaling factors. 

\subsection{Asymmetric FP3 and FP4 Data Types} \label{sec:asym_fp_type}
The basic floating-point formats contain a redundant quantization level due to the sign-magnitude representation that has both $+0$ and $-0$. We propose to replace this redundant zero with another \textit{special value} to fully utilize the available quantization levels and introduce additional asymmetry. 
We first use the basic \texttt{FP3} format to derive our custom 3-bit data type, and then extend our idea to 4-bit precision.

    \begin{table} [t]
      \centering
      
      \setlength{\tabcolsep}{4.5pt}
      \renewcommand{\arraystretch}{1.25}
      \footnotesize
      \caption{Comparison between \textit{BitMoD} and SOTA co-design frameworks for quantized LLM acceleration.}
      \vspace{-3pt}
      \begin{threeparttable}
        \begin{tabular}{ c c c c c }
            \Xhline{0.3ex}
              \multirow{2}{*}{\textbf{Framework}} & \multirow{1}{*}{\textbf{Per-group}} & \textbf{Supported} & \textbf{Accuracy\,@} & \textbf{Hardware}  
              \\
              & \textbf{Quant?} & \textbf{Precision} & \textbf{3-bit Weight} & \textbf{Efficiency}  
              \\
            \Xhline{0.3ex}
              AWQ~\cite{awq} & Yes & \textcolor{Maroon}{\textbf{Limited}} & \textcolor{Green}{\textbf{High}} & \textcolor{Maroon}{\textbf{Low}} 
              \\ \hline
              FIGNA~\cite{figna} & No & \textcolor{Maroon}{\textbf{Limited}} & \textcolor{Maroon}{\textbf{Low}} & \textcolor{Green}{\textbf{High}} 
              \\ \hline
              ANT~\cite{ant} & No & \textcolor{Maroon}{\textbf{Limited}} & \textcolor{Maroon}{\textbf{Low}} & \textcolor{Green}{\textbf{High}} 
              \\ \hline
              OliVe~\cite{olive} & No & \textcolor{Maroon}{\textbf{Limited}} & \textcolor{orange}{\textbf{Medium}} & \textcolor{Green}{\textbf{High}} 
              \\ \hline
              Microscaling~\cite{microscaling} & Yes & \textcolor{Green}{\textbf{Many}} & \textcolor{Maroon}{\textbf{Low}} & \textcolor{orange}{\textbf{Medium}} 
              \\ \hline
              \textbf{\textit{BitMoD} (Ours)} & Yes & \textcolor{Green}{\textbf{Many}} & \textcolor{Green}{\textbf{High}} & \textcolor{Green}{\textbf{High}} 
              \\
            \Xhline{0.3ex}
        \end{tabular}
      \end{threeparttable}
      \label{tab:overall_comp}
      \vspace{-5pt}
    \end{table}
    
\vspace{4pt}
\noindent
\textbf{FP3 Extension. } 
The basic \texttt{FP3} data type contains seven distinct values $\{0, \pm1, \pm2, \pm4\}$. 
Our main idea is to extend \texttt{FP3} and allows the redundant zero to be replaced by one of some pre-defined special values. Consequently, a weight group can be quantized by the basic \texttt{FP3} data type together with a selected special value to minimize the quantization error. 
Ideally, the special values can have an arbitrary precision. But a high-precision (e.g., \texttt{FP16}) special value leads to more hardware overhead for computing, which offsets the efficiency of low-precision data types. Hence, we limit the special value to low-precision integers. Furthermore, given $N$ as the number of allowed special values, an encoding overhead of $\ceil[\big]{\text{log}N}$ bits is needed to specify which special value to be selected during computation. This selection also requires an $N$-to-1 mux in the hardware implementation. To balance the encoding overhead and hardware complexity, we set $N\,=\,4$ which only requires 2-bit encoding per group. 
    
The choice of special values will affect the resulting quantization error because it changes the set of available quantization values. As discussed in Section~\ref{sec:motivation}, both asymmetry and floating-point data types are crucial for good accuracy under per-group quantization. 
Since the scaling factor and quantized values are ultimately determined by the absolute maximum value of a data type~\cite{t2c}, we establish the set of special values based on two principles. 
First, some special values should fall inside the numerical range of \texttt{FP3} to ensure that they do not alter its original absolute maximum (i.e., $4$). This is advantageous for quantizing weight groups exhibiting symmetric, Gaussian-like distribution. 
Second, some special values could fall outside the numerical range of \texttt{FP3} to introduce additional asymmetry, i.e., the absolute maximum and minimum quantization values of the extended \texttt{FP3} are different. This can benefit weight groups that exhibit asymmetric distribution.
    

    \begin{figure}
        \centering
        \includegraphics[width=1\linewidth]{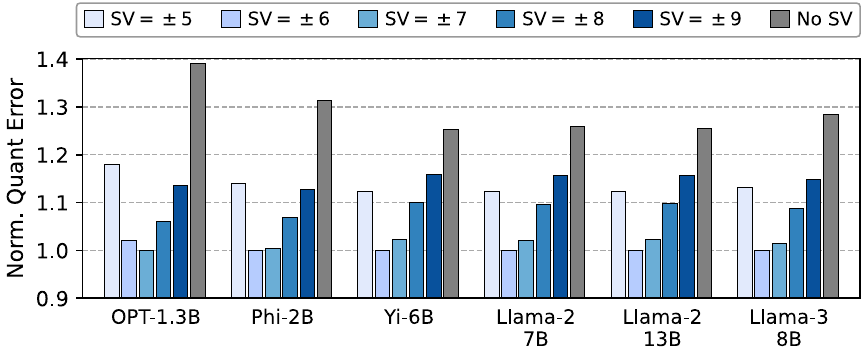}
        \vspace{-14pt}
        \caption{Normalized weight quantization error ($\downarrow$) with different special values (SV) for FP3. We use per-group quantization with a group size of 128. The special values $\pm\,6$ achieve the lowest overall quantization error, thus adopted in \textit{BitMoD}.}
        \label{fig:fp3_sv}
    \end{figure}

    \begin{table} [b]
      \centering
      \setlength{\tabcolsep}{4pt}
      \renewcommand{\arraystretch}{1.3}
      \footnotesize
      \caption{Our proposed extended resolution (ER) and extended asymmetry (EA) \texttt{FP3} and \texttt{FP4} data types.}
      \vspace{-3pt}
        \begin{tabular}{ c | c | c | c }
            \Xhline{0.3ex}
              Dtype & Basic Values & Extended Dtype & Special Value \\
            \hline
              \multirow{2}{*}{FP3} & \multirow{2}{*}{0, $\pm\,1$, $\pm\,2$, $\pm\,4$} & \multirow{1}{*}{FP3-ER} & $-3$ or $+3$ \\
            \cline{3-4}
              & & \multirow{1}{*}{FP3-EA} & $-6$ or $+6$ \\
            \hline
              \multirow{2}{*}{FP4} & \multirow{1}{*}{0, $\pm\,0.5$, $\pm\,1$, $\pm\,1.5$} & \multirow{1}{*}{FP4-ER} & $-5$ or $+5$ \\
            \cline{3-4}
              & \multirow{1}{*}{$\pm\,2$, $\pm\,3$, $\pm\,4$, $\pm\,6$} & \multirow{1}{*}{FP4-EA} & $-8$ or $+8$ \\
            \Xhline{0.3ex}
        \end{tabular}
      \label{tab:fp3_fp4_dtype}
      \vspace{-3pt}
    \end{table}
    
To satisfy the first property, the special values should be set to $\pm\,3$, 
which replace the redundant zero with $+\,3$ and $-\,3$, respectively. We call this new data type \texttt{FP3-ER} since it adds extra resolution (ER) within the range of \texttt{FP3}. 
To satisfy the second property, there are an infinite number of values that can fall outside the \texttt{FP3} range. Therefore, we determine the two remaining special values that can minimize the quantization error.
We further reduce the search space by restricting these two special values to have the same absolute value, which results in \textit{balanced} asymmetry across all weight groups. This is desirable because, although an individual weight group may prefer asymmetric quantization, a whole weight tensor usually exhibits symmetric, Gaussian-like distribution~\cite{ant, mokey}. Fig.~\ref{fig:fp3_sv} shows the normalized per-group quantization error on six LLMs when adding different special values to \texttt{FP3}. We observe that adding asymmetry significantly reduces the quantization error. In addition, the special values $\pm\,6$ have the lowest quantization error on most LLMs except for OPT-1.3B, and are therefore adopted in \textit{BitMoD}. We call the resulting new data type \texttt{FP3-EA} since it adds extra asymmetry (EA) to extend the range of \texttt{FP3}. 
    
    
\vspace{4pt}
\noindent
\textbf{FP4 Extension. } 
Similar to \texttt{FP3-ER} and \texttt{FP3-EA}, we add extra resolution and asymmetry to \texttt{FP4}. We conduct experiments to measure the effects of different \texttt{FP4} special values on the resulting quantization error, which leads to the best \texttt{FP4-ER} and \texttt{FP4-EA} that have special values $\pm\,5$ and $\pm\,8$, respectively. Table~\ref{tab:fp3_fp4_dtype} summarizes the extended \texttt{FP3} and \texttt{FP4} data types. Note that although we have fixed the four special values given that they can minimize the quantization error for the diverse set of LLMs that we evaluate, the proposed \textit{BitMoD} accelerator can flexibly accommodate other arbitrary special values that may perform well with different LLMs, which we discuss in Section~\ref{sec:bit_serial_term}.

\begin{algorithm} [t]
    \caption{Fine-grained data type adaptation.}
    \label{algo:quant}
    \setstretch{1.1}
    \DontPrintSemicolon
    \small
    
    \Input{Weight group: $W$; Quantization precision: $p$}
    \Output{Quantized weight group: $W_{qout}$\,; \\
    Selected special value: $v_{out}$}
    
    \Func{\upshape AdaptiveQuant$(\,W, \, p\,)$}{
        \tcp{Get basic and special quantization values according to Table~\ref{tab:fp3_fp4_dtype}}
        basicValues $=$ GetBasicValues$(\,p\,)$ \\
        specialValues $=$ GetSpecialValues$(\,p\,)$ \\

        \tcp{Search for the best special value}
        minError $=$ $+\infty$ \\
        \For{\upshape $v$ in specialValues} {
            quantValues $=$ basicValues $\cup$ $v$ \\
            $W_{q}$ $=$ NonLinearQuantize$(\,W,$ quantValues$\,)$ \\
            newError $=$ MeanSquareError$(\,W, \, W_{q}\,)$ \\
            \If{\upshape newError $<$ minError} {
                minError $=$ newError \\
                $W_{qout}$ $=$ $W_{q}$ \\
                $v_{out} = v$
            } 
        } 
        \Return{\upshape $W_{qout}$\,, $v_{out}$} 
    } 
\end{algorithm}



\subsection{Fine-grained Data Type Adaptation} \label{sec:dtype_adapt}
The extended \texttt{FP3} and \texttt{FP4} data types contain four special values, but every weight group can only be quantized with one special value in addition to the basic values. Therefore, we propose a \textit{fine-grained data type adaptation} strategy that quantizes every group using a different special value to minimize the quantization error, as detailed in Algo.~\ref{algo:quant}. First, the basic and special values are obtained from Table~\ref{tab:fp3_fp4_dtype} (Line 2 -- 3). We iterate through all special values and add one special value to the set of basic values in every iteration (Line 5 -- 6). Then we perform non-linear quantization (Line 7), which is commonly used in previous PTQ studies that map the floating-point tensor to a set of non-linear values (i.e., non-INT data types)~\cite{students, ant, olive}. Finally, we assign the special value that has the lowest mean-square error between the original and quantized weight (Line 8 -- 11). 

Although Algo. \ref{algo:quant} describes the quantization procedure for a single weight group, the algorithm can be vectorized on a GPU to find the best special value for all groups of a weight tensor simultaneously. 
For our implementation, the algorithm only takes ${\sim}$10~second to quantize the whole Llama-2-7B model on a single A6000 GPU.
Hence, our proposed quantization strategy exhibits high compression speed and efficiency.

\subsection{Efficient Per-group Dequantization} \label{sec:dequant}
While quantization allows computing the dot product in low precision, it still requires dequantization (i.e., re-scaling) after producing the output. Specifically, Eq.~\ref{eq:quant_sym} indicates that the quantized low-precision weight should be multiplied by the scaling factor to obtain the actual floating-point weight. Per-channel quantization only needs re-scaling after producing the final output activation. Such per-channel re-scaling can be further fused into other element-wise operations such as layer-norm before writing the output activation back to memory, reducing the data transfer cost~\cite{awq, smoothquant}. However, per-group quantization must dequantize the partial sum after computing every group of dot-products because different groups have different scaling factors. Furthermore, since \textit{BitMoD} maintains the input activation at \texttt{FP16}, the group partial sum will also have floating-point formats. As a result, performing dequantization on-the-fly necessitates a floating-point pipeline, which can diminish the potential hardware efficiency gained from using low-precision weights. 



    \begin{table} [t]
      \centering
      \setlength{\tabcolsep}{4pt}
      \renewcommand{\arraystretch}{1.15}
      \footnotesize
      \caption{Wikitext-2 and C4 perplexity ($\downarrow$) under different precision for the per-group scaling factor (SF). We use \texttt{INT4-Asym} for weight quantization with a group size of 128.}
      \vspace{-3pt}
        \begin{tabular}{ c | c  c | c  c | c  c | c  c }
            \Xhline{0.3ex}
              Model & \multicolumn{2}{c|}{OPT-1.3B} & \multicolumn{2}{c|}{Phi-2B} & \multicolumn{2}{c|}{Llama-2-7B}  & \multicolumn{2}{c}{Llama-2-13B} \\
            \cline{1-9}
              SF Bits & Wiki & C4 & Wiki & C4 & Wiki & C4 & Wiki & C4 \\
            \Xhline{0.3ex}
              FP16 & 15.41 & 15.84 & 10.68 & 13.66 & 5.77 & 7.31 & 5.01 & 6.62 \\
            \hline
              INT8 & 15.41 & 15.84 & 10.68 & 13.66 & 5.77 & 7.31 & 5.01 & 6.62 
            \\
              INT6 & 15.43 & 15.84 & 10.74 & 13.71 & 5.77 & 7.31 & 5.01 & 6.62 
            \\
              INT4 & 15.52 & 15.93 & 10.76 & 13.73 & 5.77 & 7.36 & 5.03 & 6.64 
            \\
              INT2 & 18.46 & 18.61 & 15.68 & 18.32 & 8.41 & 10.63 & 6.19 & 8.12 \\
            \Xhline{0.3ex}
        \end{tabular}
      \label{tab:sf_prec}
    \end{table}
    
In order to reduce the dequantization cost, we build upon prior work VS-Quant~\cite{vs-quant}, which applies a second-level quantization that further quantizes the scaling factors to low-precision integers. Given the weight channel size $D$ and group size $G$, VS-Quant applies symmetric quantization (Eq.~\ref{eq:quant_sym}) to the $D/G$ scaling factors from the same channel, where the precision of per-group scaling factor is a design parameter. However, VS-Quant only targets small-scale neural networks and uses a small group size of 16. It is unclear how quantizing the scaling factors of a larger group will affect the accuracy of LLMs. Hence, we conduct experiments to find the best precision for per-group scaling factors. We use \texttt{INT4-Asym} quantization as an example, while other data types show the same trend. As shown in Table~\ref{tab:sf_prec}, \texttt{INT8} per-group scaling factors have no accuracy loss compared to using \texttt{FP16} scaling factors. This is expected since \texttt{INT8} can even achieve no accuracy loss for per-channel weight quantization~\cite{smoothquant, zeroq}, which has a much wider numerical range than scaling factors. Thus, we use \texttt{INT8} per-group scaling factors in \textit{BitMoD}, which allows efficient per-group dequantization in a bit-serial manner as will be described in Section~\ref{sec:bitmod_pe}.

\vspace{4pt}
\noindent
\textbf{Memory Overhead Analysis. } 
The proposed \textit{BitMoD} quantization only needs an 8-bit scaling factor and 2-bit encoding metadata to select the special value for every group. Given a large group size such as 128, which is commonly used in SOTA software-only LLM quantization studies~\cite{gptq, awq, students}, the 10-bit extra memory per group incurs practically no overhead. Furthermore, prior software-only PTQ works mainly adopt asymmetric integer quantization~\cite{gptq, awq, students}, which requires a 16-bit scaling factor and an 8-bit zero-point per group. Hence, \textit{BitMoD} exhibits lower memory overhead compared to these works.


\section{BitMoD Hardware Accelerator}
    \begin{figure} [t]
        \centering
        \includegraphics[width=1\linewidth]{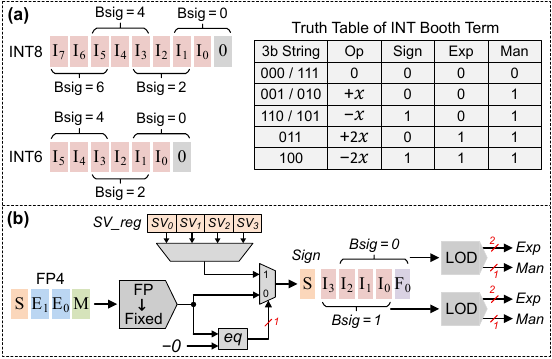}
        \vspace{-12pt}
        \caption{Unified bit-serial representation of (a) INT8, INT6, and (b) FP4. Every bit-serial term contains four parts: sign, exponent (exp), mantissa (man) and bit-significance (bsig).}
        \label{fig:bs_dtype}
    \end{figure}
    
    \begin{figure*}
        \centering
        \includegraphics[width=1\linewidth]{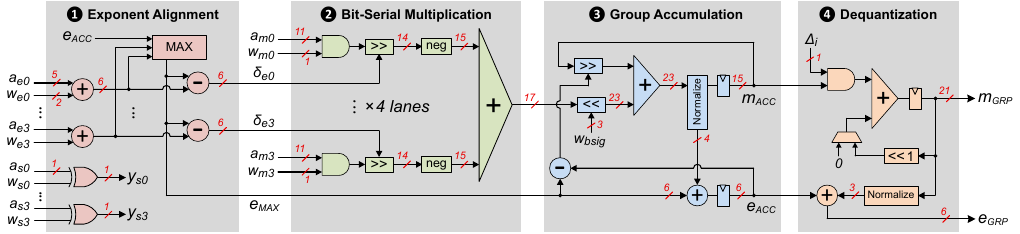}
        \vspace{-12pt}
        \caption{The microarchitecture of \textit{BitMoD} PE. Every bit-serial weight term contains 1-bit sign ($w_s$), 2-bit exponent ($w_e$), 1-bit mantissa ($w_m$), and a shared bit-significance ($w_{bsig}$). }
        \label{fig:bitmod_pe}
        \vspace{-3pt}
    \end{figure*}
    
In this section, we describe the \textit{BitMoD} hardware design, which leverages the bit-serial computing paradigm to offer a good trade-off between weight precision, model accuracy and hardware efficiency. Section~\ref{sec:bit_serial_term} develops a unified bit-serial representation of different low-precision data types supported by \textit{BitMoD}. Section~\ref{sec:bitmod_pe} details the microarchitecture of the \textit{BitMoD} processing element (PE). Finally, section~\ref{sec:bitmod_accel} presents the overall accelerator architecture.

\subsection{Unified Bit-serial Representation} \label{sec:bit_serial_term}
Prior studies have demonstrated that per-channel \texttt{INT8} weight quantization shows no accuracy loss compared to using \texttt{FP16} weights~\cite{figna, zeroq, smoothquant}. Moreover, as discussed in Section~\ref{sec:motivation}, per-group \texttt{INT6} quantization also has negligible accuracy loss. Hence, the design target of \textit{BitMoD} hardware is to support \texttt{INT8}, \texttt{INT6}, as well as the new \texttt{FP4} and \texttt{FP3} extensions in a unified architecture. However, the basic values of \texttt{FP3} and \texttt{FP4} use the floating-point format, which is not compatible with the integer representation. A naive approach is to convert all data types to \texttt{INT8}, yet this cannot improve the computational efficiency for lower-precision weights.

In order to trade-off lower weight precision for improved hardware efficiency, we propose a unified bit-serial representation, where every number is decomposed into a series of bit-serial terms, each containing four parts: sign, exponent (exp), mantissa (man), and bit-significance (bsig). The value of a bit-serial term can be expressed as:
\begin{equation}
    v_{\text{term}} = (-1)^{\text{sign}} \cdot 2^{\text{exp}} \cdot \text{man} \cdot 2^{\text{bsig}}
    \label{eq:bs_term}
\end{equation} 

    
Fig.~\ref{fig:bs_dtype} describes the bit-serial representation for different data types supported by \textit{BitMoD}. For \texttt{INT8} and \texttt{INT6}, we apply Booth encoding~\cite{booth} to decompose their binary strings into four and three 3-bit Booth strings, respectively. The Booth encoding has been widely used in prior bit-serial accelerators to speed up computation~\cite{pragmatic, laconic, fpraker}. 
Every two adjacent Booth strings have a difference of 2 in bit-significance.
The sign, exponent, and mantissa depend on the content of a Booth string, which defines the desired operation when multiplying with another operand $x$.

For the extended \texttt{FP4}, we first convert it to fixed-point values in sign-magnitude format with 1 sign bit, 4 integer bits $\{I_3, I_2, I_1, I_0\}$ to handle the largest special value $\pm\,8$ of \texttt{FP4-EA}, and 1 fraction bit $\{F_0\}$ to handle $\pm\,0.5$ and $\pm\,1.5$ of the basic \texttt{FP4} values. 
The fixed-point value is then compared with the redundant negative zero. If the comparison result is equal, the negative zero will be replaced by the assigned special value (\textit{SV}) of a particular weight group. The four allowed special values are stored in a register file (\textit{SV\_reg}), which only requires one-time programming before deploying an LLM.
To obtain the bit-serial term, we observe that all values of the extended \texttt{FP4} in Table~\ref{tab:fp3_fp4_dtype} contain at most two `1' bits after converting to the fixed-point format. Hence, we use the simple leading-one detector (\textit{LOD}) to get two bit-serial terms from the first four bits $\{I_3, I_2, I_1, I_0\}$ and last four bits $\{I_2, I_1, I_0, F_0\}$, respectively. Finally, since the extended \texttt{FP3} values are a subset of \texttt{FP4}, it can be decoded into two bit-serial terms using the same hardware. 
Note that the bit-serial decoder is not limited to support the special values shown in Table~\ref{tab:fp3_fp4_dtype}. The special value register file can be programmed with other special values as needed, and the number of decoded bit-serial terms can be minimized with simple modification to the decoder. For example, the special value $7$ can be expressed as two bit-serial terms $2^3$ and $-\,2^0$ instead of using a leading-one detector that emits three bit-serial terms.

    \begin{figure} [t]
        \centering
        \includegraphics[width=1\linewidth]{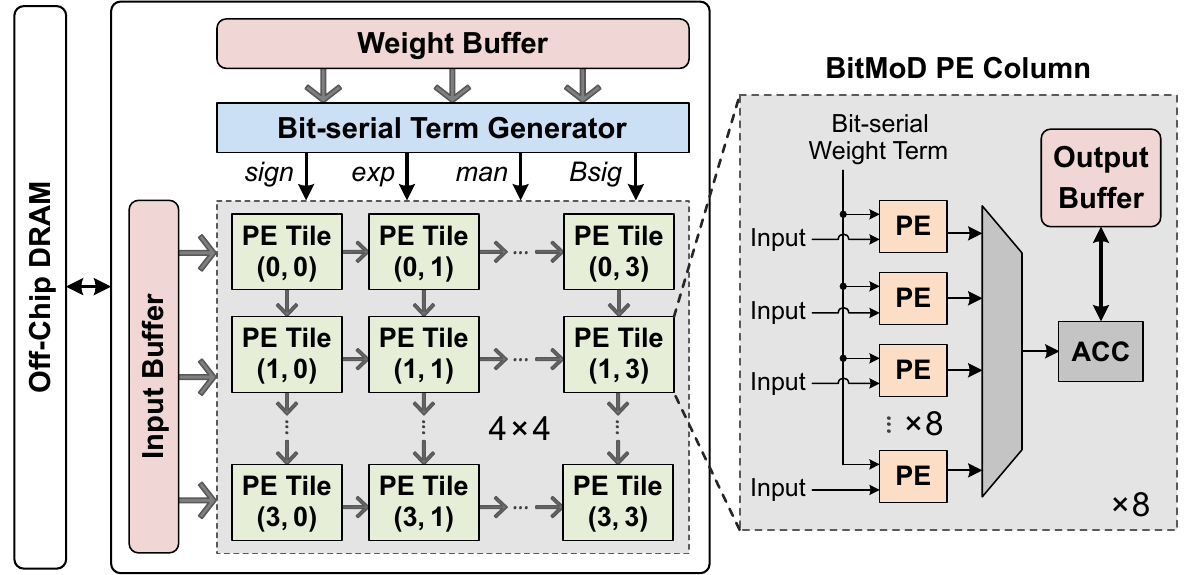}
        \vspace{-12pt}
        \caption{\textit{BitMoD} accelerator architecture.}
        \label{fig:bitmod_acc}
        \vspace{-5pt}
    \end{figure}

    \begin{table*} [t]
        \centering
        \setlength{\tabcolsep}{5.5pt}
        \renewcommand{\arraystretch}{1.1}
        \footnotesize      
        \caption{Wikitext-2 and C4 perplexity ($\downarrow$) using different data types under per-group weight quantization. }
        \vspace{-3pt}
        \begin{threeparttable}
        \begin{tabular}{ c c  c c  c c  c c c c  c c  c c |  c }
            \Xhline{0.3ex}
              \multirow{2}{*}{\textbf{Precision}} & \multirow{2}{*}{\textbf{Datatype\tnote{*}}} & \multicolumn{2}{c}{\textbf{OPT-1.3B}} & \multicolumn{2}{c}{\textbf{Phi-2B}} & \multicolumn{2}{c}{\textbf{Yi-6B}} & \multicolumn{2}{c}{\textbf{Llama-2-7B}} & \multicolumn{2}{c}{\textbf{Llama-2-13B}} & \multicolumn{2}{c|}{\textbf{Llama-3-8B}}  & \multirow{2}{*}{\textbf{Mean\,$\Delta$PPL}} 
            \\
              & & Wiki & C4 & Wiki & C4 & Wiki & C4 & Wiki & C4 & Wiki & C4 & Wiki & C4 & 
            \\
            \Xhline{0.3ex}
              16-bit & FP16 & 14.62 & 14.72 & 9.71 & 12.74 & 5.84 & 8.91 & 5.47 & 6.97 & 4.88 & 6.47 & 6.13 & 8.88 & 0
            \\
            \hline
              \multirow{5}{*}{4-bit} & ANT & 16.23 & 16.08 & 11.23 & 14.31 & 6.87 & 10.95 & 6.09 & 7.71 & 5.31 & 6.91 & 7.58 & 11.04 & 1.23
              \\
              & OliVe & 15.38 & 15.82 & 10.49 & \textbf{13.51} & 6.55 & 9.84 & 5.91 & 7.34 & 5.13 & 6.74 & 6.89 & 9.91 & 0.68
              \\
              & MX-FP4 & 15.39 & 15.81 & 10.72 & 13.72 & 6.62 & 10.24 & 5.82 & 7.39 & 5.11 & 6.71 & 7.04 & 10.13 & 0.79
              \\
              & INT4-Asym & 15.41 & 15.74 & 10.67 & 13.65 & 6.32 & 9.69 & 5.77 & 7.31 & \textbf{5.01} & 6.62 & 6.84 & 9.79
              & 0.62
              \\
              & \textit{BitMoD} & \textbf{14.89} & \textbf{15.29} & \textbf{10.48} & 13.53 & \textbf{6.23} & \textbf{9.58} & \textbf{5.72} & \textbf{7.26} & \textbf{5.01} & \textbf{6.61} & \textbf{6.73} & \textbf{9.66} & \textbf{0.48}
              \\
            \hline
              \multirow{5}{*}{3-bit} & ANT & 340.6 & 332.9 & 15.57 & 18.35 & 9.01 & 14.32 & 8.51 & 10.28 & 6.40 & 7.98 & 15.22 & 17.56 & 57.61
              \\
              & OliVe & 76.79 & 59.63 & 14.93 & 17.76 & 32.42 & 66.02 & 9.13 & 12.04 & 8.69 & 12.43 & 26.76 & 46.39 & 23.14
              \\
              & MX-FP3 & 1E+3 & 771.6 & 17.89 & 20.37 & 15.41 & 21.97 & 8.86 & 11.99 & 7.19 & 9.13 & 23.82 & 31.39 & 152.8
              \\
              & INT3-Asym & 139.4 & 144.9 & 13.92 & 16.79 & 8.66 & 13.33 & 7.08 & 9.29 & 5.64 & 7.35 & 13.26 & 17.80 & 24.34
              \\
              & \textit{BitMoD} & \textbf{22.67} & \textbf{20.47} & \textbf{12.91} & \textbf{15.69} & \textbf{7.66} & \textbf{11.98} & \textbf{6.55} & \textbf{8.36} & \textbf{5.50} & \textbf{7.18} & \textbf{8.96} & \textbf{12.82} & \textbf{2.94}
              \\
            \Xhline{0.3ex}
        \end{tabular}

        \begin{tablenotes}
          \item[\textbf{*}] All quantization data types use per-group quantization. The MX data type uses a group size of 32 following the standard in \cite{mx-format}, while other data types use a group size of 128. The perplexity of MX degrades when using a larger group size. 

        \end{tablenotes}
        \end{threeparttable}
        \label{tab:gen_task_ppl}      
        \vspace{-5pt}
      \end{table*}  

\subsection{BitMoD Processing Element} \label{sec:bitmod_pe}
While \textit{BitMoD} is able to quantize weight to low precision, activation still remains in \texttt{FP16}. To address this challenge, we propose a mixed-precision bit-serial PE as shown in Fig.~\ref{fig:bitmod_pe}.
In every cycle, the PE performs a 4-way dot product between four bit-serial weight terms~($w$) and four \texttt{FP16} activations~($a$). Step~\protect\circled{\small1} first aligns the sum of exponents ($a_e + w_e$) to compute the delta exponent ($\delta_e$). It also generates the sign ($y_s$) of every product between a weight term and activation. Step~\protect\circled{\small2} performs the bit-serial multiplication between the 1-bit weight mantissa ($w_m$) and 11-bit activation mantissa ($a_m$) including the hidden bit. The multiplication result is aligned by a right-shifter that is controlled by the delta exponent. We reserve 3 extra bits in the shifter result to account for rounding to the nearest even as suggested by Awad \textit{et al.}~\cite{fpraker}. The bit-serial dot product of the mantissa is then computed using an adder tree. After producing the bit-serial dot product, Step~\protect\circled{\small3} performs accumulation by first multiplying the dot product with the weight bit-significance ($W_{bsig}$), followed by adding with the accumulator mantissa ($m_{ACC}$). The accumulated mantissa is then normalized to update the accumulator exponent ($e_{ACC}$). Since \textit{BitMoD} adopts per-group quantization, the accumulated group partial sum must be dequantized on the fly. To reduce this hardware cost, Step~\protect\circled{\small4} performs dequantization in a bit-serial manner. Specifically, the accumulator mantissa is multiplied by one bit of the group scaling factor ($\Delta_i$) in every cycle, followed by shift-and-add to obtain the exponent ($e_{GRP}$) and mantissa ($m_{GRP}$) of the dequantized partial sum.  
    
One concern of the bit-serial dequantization is whether it will take more cycles than the normal dot product stage and cause potential pipeline stalling. As discussed in Section~\ref{sec:dequant}, the per-group scaling factor has 8 bits, which requires 8 cycles for dequantization. On the other hand, even the lowest-precision data type \texttt{FP3} in \textit{BitMoD} requires two cycles to process two bit-serial terms. Given the PE dot-product size of 4 and a commonly used group size of 128, the group dot-product stage takes $128 / 4 \times 2 = 64$ cycles to complete. Therefore, the proposed bit-serial dequantization will never stall the computing pipeline. Furthermore, since \texttt{INT6} and the extended \texttt{FP4}/\texttt{FP3} data types contain three and two bit-serial terms, the proposed \textit{BitMoD} PE is able to compute 4 multiply-accumulate operations in 3 and 2 cycles, respectively. Compared to the normal \texttt{FP16} multiply-accumulate hardware, \textit{BitMoD} achieves a throughput improvement of $1.33\times$ and $2\times$ for \texttt{INT6} and \texttt{FP4}/\texttt{FP3} data types, respectively. In fact, as will be evaluated in Section~\ref{sec:hardware_eval}, the \textit{BitMoD} PE consumes $24\%$ less area than an \texttt{FP16} PE, thus is able to provide even higher throughput under an iso-compute area constraint. 

Besides matrix multiplication between weights and activations, LLMs also contain the self-attention layer, which involves two matrix multiplication operations between three activation tensors, i.e., query, key, and value. Given that the \textit{BitMoD} PE only maintains one activation tensor in \texttt{FP16}, the other two activation tensors need to be low-precision integers. Fortunately, prior works have demonstrated that the key and value tensors are very amenable to quantization due to the softmax normalization inside self-attention, and can be safely quantized to \texttt{INT8} and even \texttt{INT4} with negligible accuracy loss~\cite{flexgen, atom, smoothquant}. Hence, \textit{BitMoD} can accommodate the self-attention layer with the proposed bit-serial PE by quantizing the key and value tensors to low-precision integers.

\subsection{BitMoD Accelerator} \label{sec:bitmod_accel}
Fig. \ref{fig:bitmod_acc} shows the overall architecture of the \textit{BitMoD} accelerator. The input and weight buffers are banked to provide adequate bandwidth for the access from PEs. The bit-serial term generator receives the weight data and decomposes them into bit-serial terms as discussed in Section~\ref{sec:bit_serial_term}. The main PE array contains $4 \times 4$ tiles connected in a systolic manner. Every PE tile has 8 rows $\times$ 8 columns and adopts an output-stationary dataflow. The bit-serial weight term is broadcast to the whole PE column, while the input is broadcast to the whole PE row. This allows \textit{BitMoD} to exploit data reuse through both weight-sharing and input-sharing. Every PE column contains a local output buffer and an accumulator, which is used to accumulate the per-group partial sum from a PE to obtain the final per-channel output activation. Since processing a weight group takes many cycles, there is enough time to drain the whole PE column using only one shared accumulator to amortize the hardware cost.

\section{Evaluation} \label{sec:evaluation}

\subsection{Experimental Methodology} \label{sec:methodology}
      
      \begin{table*}
        \centering
        \scriptsize
        \setlength{\tabcolsep}{2.5pt}
        \renewcommand{\arraystretch}{1.15}
        \caption{Accuracy ($\uparrow$) of discriminative tasks using different data types under per-group weight quantization. }
        \vspace{-3pt}
        \begin{tabular}{ c c  c c c  c c c  c c c  c c c  c c c  c c c |  c }
            \Xhline{0.3ex}
              \multirow{2}{*}{\textbf{Precision}} & \multirow{2}{*}{\textbf{Datatype}} & \multicolumn{3}{c}{\textbf{OPT-1.3B}} & \multicolumn{3}{c}{\textbf{Phi-2B}} & \multicolumn{3}{c}{\textbf{Yi-6B}} & \multicolumn{3}{c}{\textbf{Llama-2-7B}} & \multicolumn{3}{c}{\textbf{Llama-2-13B}} & \multicolumn{3}{c|}{\textbf{Llama-3-8B}} & \multirow{2}{*}{\textbf{Mean\,$\Delta$Acc}}
              \\
              & & \;Hella & Wino & Piqa\; & \;Hella & Wino & Piqa\; & \;Hella & Wino & Piqa\; & \;Hella & Wino & Piqa\; & \;Hella & Wino & Piqa\; & \;Hella & Wino & Piqa\;  
              \\
            \Xhline{0.3ex}
              16-bit & FP16 & \;53.72 & 59.43 & 72.41\; & \;73.74 & 75.77 & 79.22\; & \;74.96 & 70.72 & 78.78\; & \;75.98 & 69.06 & 79.11\; & \;79.39 & 72.38 & 80.5\; & \;79.18 & 72.85 & 80.74\; & 0 
              \\
            \hline
              \multirow{2}{*}{4-bit} & INT4-Asym & \;52.31 & \textbf{59.35} & 71.05\; & \;72.29 & 75.14 & 78.4\; & \;73.91 & \textbf{70.51} & 77.64\; & \;75.29 & \textbf{68.74} & 78.22\; & \;\textbf{78.76} & \textbf{72.45} & 80.2\; & \;78.07 & \textbf{73.24} & 79.76\; & -0.71
              \\
              & \textit{BitMoD} & \;\textbf{53.03} & 59.12 & \textbf{71.49}\; & \;\textbf{72.51} & \textbf{77.58} & \textbf{79.48}\; & \;\textbf{73.98} & 70.09 & \textbf{78.35}\; & \;\textbf{75.43} & 68.19 & \textbf{78.45}\; & \;78.41 & 72.14 & \textbf{80.42}\; & \;\textbf{78.49} & 73.09 & \textbf{79.98}\; & \textbf{-0.42}
              \\
            \hline
              \multirow{2}{*}{3-bit} & INT3-Asym & \;38.98 & 55.01 & 64.25\; & \;67.75 & 71.74 & 77.48\; & \;\textbf{71.3} & 67.32 & \textbf{76.71}\; & \;71.87 & \textbf{66.46} & 76.66\; & \;76.58 & 69.61 & 78.94\; & \;68.56 & 66.61 & 75.03\; & -4.84
              \\
              & \textit{BitMoD} & \;\textbf{49.16} & \textbf{58.09} & \textbf{68.88}\; & \;\textbf{70.16} & \textbf{75.22} & \textbf{78.18}\; & \;70.72 & \textbf{67.72} & 76.28\; & \;\textbf{72.68} & 66.22 & \textbf{77.53}\; & \;\textbf{76.79} & \textbf{72.37} & \textbf{79.22}\; & \;\textbf{73.56} & \textbf{70.32} & \textbf{77.91}\; & \textbf{-2.61}
              \\
            \Xhline{0.3ex}
        \end{tabular}
      \label{tab:dis_task_acc}
      \vspace{-5pt}
    \end{table*}
    
\noindent
\textbf{LLM Benchmarks. } 
For evaluation, we choose six representative LLMs with diverse model sizes, including \text{OPT-1.3B}~\cite{opt}, \text{Phi-2B}~\cite{phi-2b}, \text{Yi-6B}~\cite{yi-6b}, \text{Llama-2-7B}, \text{Llama-2-13B}~\cite{llama-2}, and \text{Llama-3-8B}~\cite{llama-3}. We obtain the pre-trained models from the HuggingFace repository, and implement the proposed \textit{BitMoD} quantization framework in PyTorch. To evaluate the effects of quantization on the resulting model performance, we consider both discriminative and generative tasks. For discriminative tasks, we evaluate three benchmarks: HellaSwag~\cite{hellaswag}, WinoGrande~\cite{winogrande}, and Piqa~\cite{piqa} under the zero-shot setting using LM-Evaluation-Harness~\cite{lm-eval}. For generative tasks, we choose Wikitext-2~\cite{wikitext} and C4~\cite{c4} datasets and evaluate the perplexity following the methodology in prior quantization works~\cite{gptq, awq, quarot, omniquant}.

\vspace{4pt}
\noindent
\textbf{Quantization Data Types. } We compare the model accuracy of \textit{BitMoD} with four baseline quantization data types: 
\begin{itemize}
  \item ANT~\cite{ant}, which adaptively uses different data types to quantize different tensors at the per-channel granularity.
  \item OliVe~\cite{olive}, that introduces an outlier-victim pair encoding mechanism, which sacrifices the normal value (i.e., victim) adjacent to the outlier to accommodate the important outlier value. 
  \item Microscaling (MX)~\cite{microscaling}, which groups 32 low-precision FP weights with an extra 8-bit shared exponent.
  \item Per-group asymmetric integer quantization, which is commonly adopted in prior software quantization methods. 
\end{itemize}
To ensure a fair comparison with \textit{BitMoD}, we only apply the data types of ANT, OliVe, and MX to LLM weight quantization while maintaining activations in \texttt{FP16}. 
Despite that the original ANT and OliVe frameworks only support per-channel quantization due to the absence of dedicated dequantization hardware, we have extended their algorithms for per-group quantization. This allows to compare the model accuracy purely based on the employed quantization data types. 
While we mainly focus on weight quantization in our evaluation, Section~\ref{sec:comb_sota_quant} demonstrates that \textit{BitMoD} can be combined with SOTA activation quantization scheme, which offers the potential to reduce activation memory as well. 

    \begin{table} [b]
      \vspace{-3pt}
      \centering
      \footnotesize
      \setlength{\tabcolsep}{4pt}
      \renewcommand{\arraystretch}{1.15}
      \caption{Wikitext-2 and C4 perplexity ($\downarrow$) when quantizing Llama weights using different data types. We use per-group weight quantization with a group size of 128.}
      \vspace{-3pt}
        \begin{tabular}{ c c  c c  c c  c c }
            \Xhline{0.3ex}
              \multirow{2}{*}{\textbf{Precision}} & \multirow{2}{*}{\textbf{Datatype}} & \multicolumn{2}{c}{\textbf{Llama-2-7B}} & \multicolumn{2}{c}{\textbf{Llama-2-13B}} & \multicolumn{2}{c}{\textbf{Llama-3-8B}}
              \\
              & & Wiki & C4 & \,Wiki & C4\, & Wiki & C4 
              \\
            \Xhline{0.3ex}
              \multirow{4}{*}{4-bit} & FP4 & 5.77 & 7.32 & \,5.05 & 6.66\, & 6.86 & 9.85 
              \\
              & FP4-ER & 5.74 & 7.28 & \,5.03 & 6.63\, & 6.76 & 9.71 
              \\
              & FP4-EA & 5.81 & 7.30 & \,5.08 & 6.65\, & 6.83 & 9.79 
              \\
              & \textit{BitMoD} & \textbf{5.72} & \textbf{7.26} & \,\textbf{5.01} & \textbf{6.61}\, & \textbf{6.73} & \textbf{9.66} 
              \\
            \hline
              \multirow{4}{*}{3-bit} & FP3 & 7.51 & 10.28 & \,5.90 & 7.58\, & 15.22 & 19.87 
              \\
              & FP3-ER & 7.18 & 9.71 & \,5.66 & 7.33\, & 13.43 & 17.56 
              \\
              & FP3-EA & 6.61 & 8.45 & \,5.54 & 7.23\, & 9.06 & 12.97
              \\
              & \textit{BitMoD} & \textbf{6.55} & \textbf{8.36} & \,\textbf{5.50} & \textbf{7.18}\, & \textbf{8.96} & \textbf{12.82} 
              \\
            \Xhline{0.3ex}
        \end{tabular}
        \vspace{-2pt}
      \label{tab:dtype_ablation}
    \end{table}
    
In addition to co-design works, we show that \textit{BitMoD} can be seamlessly integrated into existing software-only quantization optimizations to further reduce the memory footprint or achieve better model performance. We combine \textit{BitMoD} with three software quantization methods:
\begin{itemize}
  \item SmoothQuant~\cite{smoothquant}, which targets both weight and activation quantization in \texttt{INT8}. It addresses the quantization difficulty of large activation magnitude by partially migrating it to weights. 
  \item AWQ~\cite{awq}, which employs activation-aware weight quantization to protect the salient weight channels corresponding to larger activation magnitudes. 
  \item OmniQuant~\cite{omniquant}, which modulates the outlier weight values by optimizing the clipping threshold through block-wise fine-tuning.
\end{itemize}
To integrate \textit{BitMoD} with these software-only methods, we replace their original weight quantizers that use integer data types with the extended \texttt{FP4} and \texttt{FP3} data types of \textit{BitMoD}.

\vspace{4pt}
\noindent
\textbf{Accelerator Baselines. }
To evaluate the hardware performance and energy efficiency, we compare \textit{BitMoD} with a baseline accelerator that supports \texttt{FP16} models and uses an \texttt{FP16} multiply-accumulate PE instead of the proposed bit-serial PE. We also compare \textit{BitMoD} with ANT and OliVe, which design custom decoders to support multiple data types in a unified systolic array. 
We evaluate the performance in LLMs with a batch size of 1 and an input sequence length of 256, catering for edge use cases as in prior work~\cite{awq}.

\vspace{4pt}
\noindent
\textbf{Hardware Implementation. } 
We implement the accelerator of \textit{BitMoD} at RTL-level using SystemVerilog and verify the functionality of each component via RTL simulation. We use Synopsys Design Compiler to synthesize \textit{BitMoD} in TSMC 28nm technology to report the area and power. For end-to-end performance evaluation, we implement a cycle-level simulator, where the accelerator timing and energy parameters are set based on the RTL synthesis results.
The DRAM power is calculated based on the DDR4 model from DRAMSim3 \cite{dramsim}. All accelerators are evaluated under an iso-compute area constraint, and equipped with 512 KB activation buffer and 512 KB weight buffer, which are modelled with CACTI \cite{cacti}. 

    \begin{table} [b]
    \vspace{-3pt}
      \centering
      \footnotesize
      \renewcommand{\arraystretch}{1.15}
      \setlength{\tabcolsep}{2.5pt}
      \caption{Wikitext-2 and C4 perplexity ($\downarrow$) when using different special values for FP3.}
      \vspace{-3pt}
        \begin{tabular}{ c  c c  c c  c c  c c }
            \Xhline{0.3ex}
              \textbf{Special} & \multicolumn{2}{c}{\textbf{OPT-1.3B}} & \multicolumn{2}{c}{\textbf{Phi-2B}} & 
              \multicolumn{2}{c}{\textbf{Llama-2-7B}} & \multicolumn{2}{c}{\textbf{Llama-3-8B}}
              \\
              \textbf{Values} & \;Wiki & C4\; & \;Wiki & C4\; & \;Wiki & C4\; & \;Wiki & C4 
              \\
            \Xhline{0.3ex}
              $\{\pm\,5, \, \pm\,6\}$ & \;23.39 & \textbf{20.12}\; & \;13.02 & 15.84\; & \;6.61 & 8.48\; & \;9.09 & 13.81
              \\
              $\{\pm\,3, \, \pm\,5\}$ & \;35.54 & 37.65\; & \;13.41 & 16.29\; & \;6.68 & 8.73\; & \;10.32 & 14.48
              \\
              $\{\pm\,3, \, \pm\,6\}$ & \;\textbf{22.67} & 20.47\; & \;\textbf{12.91} & \textbf{15.69}\; & \;\textbf{6.55} & \textbf{8.36}\; & \;\textbf{8.96} & \textbf{12.82}
              \\
            \Xhline{0.3ex}
        \end{tabular}
        \vspace{-2pt}
        \label{tab:sv_ablation}
    \end{table}

    \begin{figure*}
        \centering
        \includegraphics[width=1\linewidth]{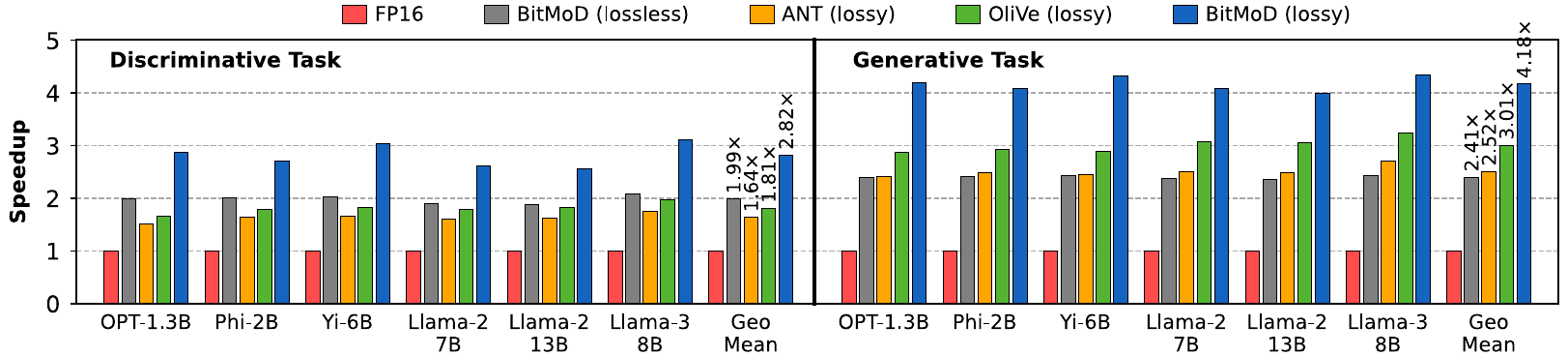}
        \vspace{-12pt}
        \caption{Speedup ($\uparrow$) of different accelerators.}
        \label{fig:speedup}
        \vspace{3pt}
    \end{figure*}

    \begin{figure*}
        \centering
        \includegraphics[width=1\linewidth]{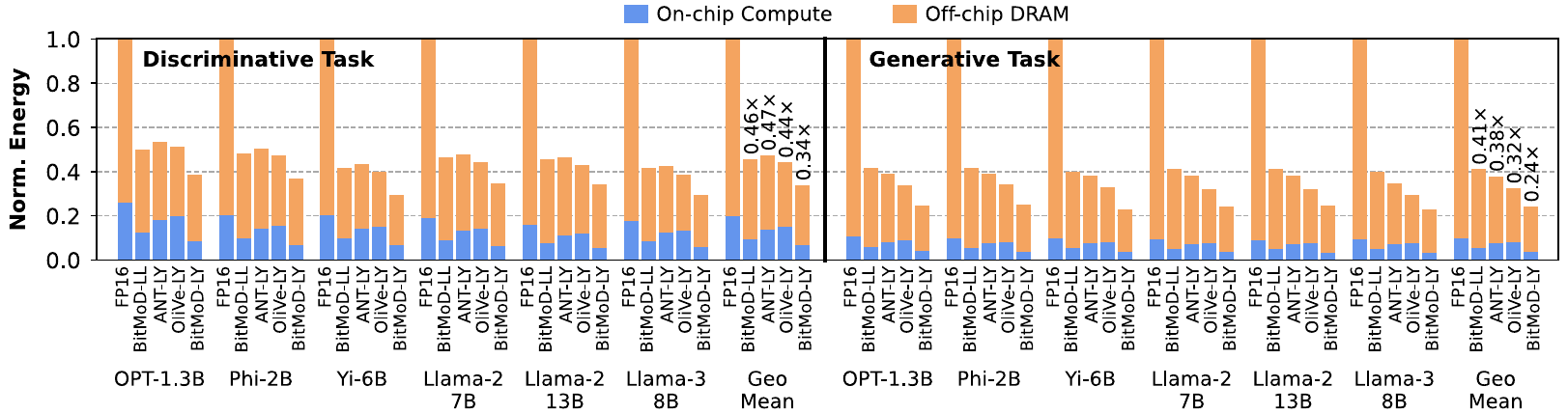}
        \vspace{-12pt}
        \caption{Energy consumption ($\downarrow$) of different accelerators. ``LL'' and ``LY'' stand for `lossless'' and `lossy'', respectively.}
        \label{fig:energy}
    \end{figure*}

\subsection{Accuracy Comparison of Different Data Types} \label{sec:accuracy}
\noindent
\textbf{Generative Tasks. } 
Table \ref{tab:gen_task_ppl} details the perplexity of applying different PTQ data types at 4-bit and 3-bit weight precision. 
For 4-bit and 3-bit weight quantization, \textit{BitMoD} achieves ${<0.5}$ and ${<3}$ perplexity loss on average compared to the \texttt{FP16} baseline models, respectively.
Although ANT, OliVe, and MX are able to maintain acceptable perplexity at 4-bit precision, they experience significant degradation in perplexity when quantizing weights to 3 bits. 
OliVe, which is designed to handle outlier values under per-channel quantization, finds its advantages diminished since the impact of outliers can be significantly mitigated through per-group quantization as described in Section~\ref{sec:motivation}. MX employs the basic \texttt{FP4} and \texttt{FP3} without exploring the potential of their redundant zero, leading to worse perplexity than \texttt{INT-Asym} that fully utilizes the available quantization levels while supporting asymmetry.
On the contrary, \textit{BitMoD} consistently outperforms asymmetric integer quantization at per-group granularity, and the benefits are more pronounced at 3-bit precision. This demonstrates that the combination of asymmetry and floating-point data types in \textit{BitMoD} can significantly reduce the quantization error. 
    
\vspace{4pt}
\noindent
\textbf{Discriminative Tasks. } 
Table \ref{tab:dis_task_acc} compares the model accuracy of discriminative tasks when employing \textit{BitMoD} and the baseline asymmetric integer quantization 
at per-group granularity. \textit{BitMoD} achieves better or comparable accuracy than asymmetric integer quantization. At 4-bit precision, \textit{BitMoD} has ${<0.5\%}$ accuracy loss on average compared to the baseline \texttt{FP16} models. Moreover, on average, \textit{BitMoD} achieves a big improvement of $2.2\%$ in model accuracy compared to the asymmetric integer quantization that is widely used in SOTA software quantization methods.
    
\vspace{4pt}
\noindent
\textbf{\textit{BitMoD} Data Type Ablation. } As discussed in Section~\ref{sec:asym_fp_type}, \textit{BitMoD} introduces new data types by adding extra resolution and asymmetry to \texttt{FP3} and \texttt{FP4}. We analyze the effects of these different data types on the perplexity of three studied Llama models. As shown in Table~\ref{tab:dtype_ablation}, the \textit{BitMoD} data types with both extra resolution and asymmetry achieve the best perplexity. Compared to the basic \texttt{FP4} data type, \texttt{FP4-ER} shows a greater improvement in perplexity than \texttt{FP4-EA}. This is because at 4-bit, the basic \texttt{FP4} still has enough quantization levels to quantize a weight group with asymmetric distribution, and adding extra resolution can better reduce the quantization error. On the contrary, for 3-bit precision, \texttt{FP3-EA} achieves much better perplexity than \texttt{FP3-ER}. Given fewer quantization levels at 3-bit, the extra asymmetry introduced by \texttt{FP3-EA} has a larger impact when accounting for weight groups with asymmetric distribution. 

    \begin{table} [b]
      \centering
      \vspace{-3pt}
      \footnotesize
      \setlength{\tabcolsep}{2.5pt}
      \renewcommand{\arraystretch}{1.25}
      \caption{Area and power consumption per tile of the baseline FP16 accelerator vs. \textit{BitMoD} at 1 GHz frequency.}
      \vspace{-3pt}
        \begin{tabular}{ c | c | c c c | c c c }
            \Xhline{0.3ex}
              \multirow{2}{*}{} & Number & \multicolumn{3}{c|}{Area ($\mu$m$^2$)} & \multicolumn{3}{c}{Power (mW)} 
              \\
              & of PEs & PE\,Array & Encoder & Total & PE\,Array & Encoder & Total 
              \\
            \hline
              Baseline & $6 \times 8$ & 95,498 & -- & 95,498 & 36.96 & -- & 36.96 \\
              \textit{BitMoD} & $8 \times 8$ & 97,090 & 2,419 & 99,509 & 37.5 & 1.86 & 39.36 \\
            \Xhline{0.3ex}
        \end{tabular}
        \vspace{-3pt}
      \label{tab:pe_area_power}
    \end{table}
    
\vspace{4pt}
\noindent
\textbf{\textit{BitMoD} Special Value Ablation. } 
The \textit{BitMoD} PE can flexibly support different special values. We evaluate two other potential combinations of special values for \texttt{FP3}: $\{\pm\,3, \, \pm\,5\}$ and $\{\pm\,5, \, \pm\,6\}$. Table~\ref{tab:sv_ablation} shows the resulting Wikitext and C4 perplexity. 
The adopted special values in \textit{BitMoD}, i.e., $\{\pm\,3, \, \pm\,6\}$, achieve the lowest perplexity on average. The special value combination $\{\pm\,5, \, \pm\,6\}$ only introduces extra asymmetry to the basic \texttt{FP3}. However, many weight groups can exhibit symmetric distribution, which prefers a symmetric data type with the same absolute maximum and minimum values. On the other hand, the special values $\pm\,5$ have a higher quantization error than $\pm\,6$ as described in Section~\ref{sec:asym_fp_type}, leading to worse perplexity when combined with $\pm\,3$. 

\subsection{Accelerator Performance} \label{sec:hardware_eval}
For accelerator evaluation, we consider two configurations for the \textit{BitMoD} accelerator based on the resulting model accuracy: (1) \textit{Lossless}, where the weight precision is \texttt{INT6} given its near-zero accuracy loss under per-group quantization. We compare this configuration with the baseline FP16 accelerator. (2) \textit{Lossy}, where the weight can be quantized to 4-bit for discriminative tasks and 3-bit for generative tasks while maintaining good model performance. We compare this configuration with ANT and OliVe.

\vspace{4pt}
\noindent
\textbf{Tile Area and Power. }
Table~\ref{tab:pe_area_power} shows the PE tile area and power breakdown of \textit{BitMoD} and the baseline FP16 accelerator at 1 GHz frequency. The unified bit-serial representation allows \textit{BitMoD} to support different weight data types with low hardware cost. As a result, the \textit{BitMoD} PE is $24\%$ smaller than the baseline PE, which allows to fit more \textit{BitMoD} PEs within the same compute area. Furthermore, the bit-serial term encoder has a tiny hardware overhead and only accounts for $2.5\%$ of the PE array area.

\vspace{4pt}
\noindent
\textbf{Performance. }
Fig. \ref{fig:speedup} presents the hardware performance normalized to the baseline FP16 accelerator for discriminative and generative tasks. \textit{BitMoD} achieves the best performance under both lossless and lossy quantization. The performance gain of \textit{BitMoD} mainly comes from its careful algorithm-hardware co-design of different quantization data types. Since discriminative tasks are compute-bound and mainly involve matrix-matrix multiplications, the higher throughput of \textit{BitMoD} PE leads to better performance than the baseline PE. In contrast, OliVe requires more complicated PEs with significant hardware overhead to accommodate the outliers, which have a much wider numerical range (e.g., $\{24, ..., 192\}$ at 4-bit precision). In comparison, the \textit{BitMoD} data type has a small value range and can be efficiently processed with the proposed unified bit-serial representation. Regarding memory-bound generative tasks, \textit{BitMoD} can quantize LLM weights to very low precision such as 3-bit, which offers significant memory saving while maintaining good perplexity. On the contrary, ANT and OliVe do not natively support per-group quantization due to the lack of dedicated dequantization hardware, and cannot maintain acceptable model quality under per-channel quantization using 3-bit weight precision. Therefore, they must adopt a higher weight precision to compensate for the significant degradation in perplexity. Overall, the lossless \textit{BitMoD} achieves $1.99\times$ and $2.41\times$ speedup for discriminative and generative tasks, respectively compared to the baseline FP16 architecture. The lossy \textit{BitMoD} achieves $1.72\times$\;/\;$1.56\times$ and $1.66\times$\;/\;$1.39\times$ speedup for discriminative and generative tasks, respectively compared to ANT\;/\;OliVe. 

    \begin{figure}
        \centering
        \includegraphics[width=1\linewidth]{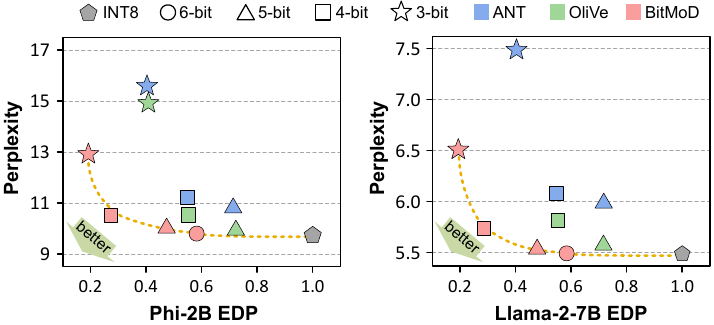}
        \vspace{-12pt}
        \caption{Wikitext-2 perplexity-EDP pareto plot for Phi-2B and Llama-2-7B.}
        \label{fig:pareto}
    \end{figure}
    
\vspace{4pt}
\noindent
\textbf{Energy Consumption. }
Fig. \ref{fig:energy} presents the normalized energy breakdown of different accelerators, where the on-chip compute energy includes both buffer and core energy. The energy saving of \textit{BitMoD} mainly comes from the reduced weight memory footprint and efficient bit-serial PE. Both ANT and OliVe require a higher weight precision than \textit{BitMoD} to maintain acceptable model quality, leading to higher DRAM energy consumption. In addition, the baseline architecture uses \texttt{FP16} weights, which is an overkill for LLMs since the simple \texttt{INT6} data type can achieve comparable accuracy under per-group weight quantization. Overall, the lossless \textit{BitMoD} achieves $2.31\times$ better energy efficiency over the baseline architecture across different tasks. The lossy \textit{BitMoD} has $1.48\times$ and $1.31\times$ better energy efficiency than ANT and OliVe, respectively.

\vspace{4pt}
\noindent
\textbf{Accuracy-Efficiency Trade-offs. }
The proposed \textit{BitMoD} can offer good trade-offs between model accuracy and hardware efficiency. To demonstrate this, we analyze the relationship between energy-delay product (EDP) and model perplexity of Phi-2B and Llama-2-7B on Wikitext-2. We compare \textit{BitMoD} with ANT and OliVe under different LLM weight precision. Fig. \ref{fig:pareto} shows the resulting perplexity-EDP relationship for the studied two LLMs and three accelerators. Note that while the 5-bit precision is not presented explicitly, \textit{BitMoD} can be easily extended to perform bit-serial \texttt{INT5} computation using its Booth encoder. Similarly, the custom data types introduced by ANT and OliVe can be extended to 5-bit precision based on their data type definition. 
As indicated in Fig. \ref{fig:pareto}, the lower left region indicates a \textit{better} trade-off between perplexity and EDP. Although ANT and OliVe propose different algorithm-hardware co-design approaches for LLM acceleration, they only leverage the per-channel quantization granularity that fails to preserve the model quality at very low precision, and they lack a unified architecture to efficiently support different data types and precision. In contrast, \textit{BitMoD} exploits new data types tailored for per-group quantization and adopts an efficient bit-serial computing paradigm to support various data types. Hence, \textit{BitMoD} can always sit on the Pareto frontier. 

    \begin{figure}
        \centering
        \includegraphics[width=1\linewidth]{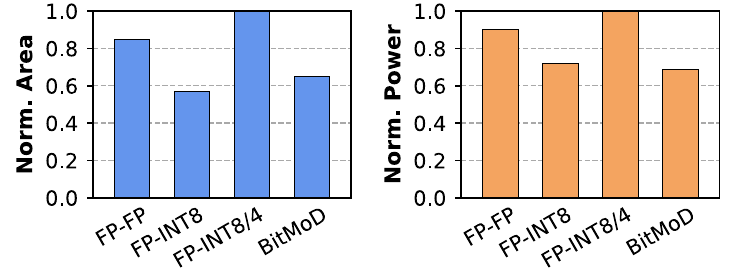}
        \vspace{-12pt}
        \caption{Normalized area and power of \textit{BitMoD} and different bit-parallel PEs.}
        \label{fig:bp_bs_comp}
    \end{figure}

\subsection{Comparison to Mixed-Precision Bit-Parallel Architecture} \label{sec:methodology}
FIGNA~\cite{figna} proposes a family of bit-parallel PEs for arithmetic between low-precision integer weight and floating-point activation. However, every FIGNA PE is designed separately and only supports one weight precision, which fails to offer a trade-off between model accuracy and hardware efficiency. We explore the possibility of FIGNA to support mixed-precision integer weights. We consider a baseline FIGNA-like PE that performs multiply-accumulate between \texttt{FP16} activation and \texttt{INT8} weight. We extend the baseline PE to support either one \texttt{FP16}-\texttt{INT8} operation or two \texttt{FP16}-\texttt{INT4} operations, which multiply the same \texttt{FP16} activation with two \texttt{INT4} weights. Fig.~\ref{fig:bp_bs_comp} compares the normalized area and power of different PEs. Although the \texttt{FP}-\texttt{INT8} PE has the smallest area, adding support for mixed weight precision incurs significant hardware overhead and leads to even higher area and power consumption than the conventional \texttt{FP}-\texttt{FP} PE. This is because a bit-parallel PE computing two \texttt{FP16}-\texttt{INT4} operations will produce two separate outputs, doubling the cost of a floating-point accumulator and output register. In contrast, the bit-serial \textit{BitMoD} PE trades-off between weight precision and latency, which requires only one accumulator and output register for any weight precision. Consequently, \textit{BitMoD} offers the highest flexibility to support variable weight precision with better efficiency compared to the decomposable bit-parallel PE.
    
    \begin{table} [t]
      \centering
      \setlength{\tabcolsep}{2pt}
      
      \renewcommand{\arraystretch}{1.15}
      \footnotesize
      \caption{Wikitext-2 and C4 perplexity ($\downarrow$) of different quantization strategies. We use per-group weight quantization with a group size of 128. For every table column, we highlight the best two perplexity results in bold.}
      \vspace{-3pt}
        \begin{tabular}{ c c c c  c c  c c  | c }
            \Xhline{0.3ex}
              \multirow{2}{*}{\textbf{Bits}} & \multirow{2}{*}{\textbf{Method}} & \multicolumn{2}{c}{\textbf{Llama-2-7B}} & \multicolumn{2}{c}{\textbf{Llama-2-13B}} & \multicolumn{2}{c}{\textbf{Llama-3-8B}} & \textbf{Mean}
              \\
              & & Wiki & C4 & \;Wiki & C4 & \;Wiki & C4\; & \textbf{$\Delta$PPL}
              \\
            \Xhline{0.3ex}
              16-bit & FP16 & 5.47 & 6.97 & \;4.88 & 6.47 & \;6.13 & 8.88\; & 0 
              \\
            \hline
              \multirow{6}{*}{4-bit} & QuaRot & 5.60 & 7.48 & \;5.00 & 6.88 & \;6.54 & 10.18\; & 0.48
              \\
              & GPTQ & 5.63 & 7.13 & \;4.99 & 6.56 & \;6.53 & 9.38\; & 0.24
              \\
              & AWQ & 5.60 & 7.12 & \;4.97 & 6.56 & \;6.54 & 9.39\; & 0.23
              \\
              & OmniQ & 5.59 & 7.12 & \;4.96 & 6.56 & \;6.57 & 9.50\; & 0.25
              \\
              & \textit{BitMoD}\;+\;AWQ & \textbf{5.59} & \textbf{7.09} & \;\textbf{4.96} & \textbf{6.55} & \;\textbf{6.50} & \textbf{9.33}\; & \textbf{0.20}
              \\
              & \textit{BitMoD}\;+\;OmniQ & \textbf{5.57} & \textbf{7.07} & \;\textbf{4.95} & \textbf{6.55} & \;\textbf{6.45} & \textbf{9.30}\; & \textbf{0.18}
              \\
            \hline
              \multirow{6}{*}{3-bit}
              & QuaRot & 6.09 & 8.44 & \;5.37 & 7.52 & \;\textbf{7.64} & 12.49\; & 1.88
              \\
              & GPTQ & 6.29 & 7.89 & \;5.42 & 7.00 & \;9.58 & 11.66\; & 1.51
              \\
              & AWQ & 6.24 & 7.81 & \;5.32 & 6.95 & \;8.22 & 11.56\; & 1.22
              \\
              & OmniQ & \textbf{6.05} & 7.76 & \;5.28 & 6.99 & \;8.33 & 12.04\; & 1.28
              \\
              & \textit{BitMoD}\;+\;AWQ & 6.07 & \textbf{7.64} & \;\textbf{5.27} & \textbf{6.88} & \;7.81 & \textbf{11.07}\; & \textbf{0.98}
              \\
              & \textit{BitMoD}\;+\;OmniQ & \textbf{5.89} & \textbf{7.59} & \;\textbf{5.21} & \textbf{6.85} & \;\textbf{7.57} & \textbf{11.05}\; & \textbf{0.89}
              \\
            \Xhline{0.3ex}
        \end{tabular}
      \label{tab:sota_quant}
      \vspace{-5pt}
    \end{table}

\subsection{Combining \textit{BitMoD} with Other Quantization Schemes} \label{sec:comb_sota_quant}
\textit{BitMoD} can be seamlessly integrated with existing software-only quantization methods by replacing their original weight quantizers that use integer data types with the extended \texttt{FP4} and \texttt{FP3} data types of \textit{BitMoD}. We demonstrate such feasibility on AWQ~\cite{awq}, OmniQuant~\cite{omniquant}, and SmoothQuant~\cite{smoothquant}.

\vspace{4pt}
\noindent
\textbf{Orthogonal to Quantization Optimization. } 
The original AWQ and OmniQuant adopt \texttt{INT-Asym} weight quantization with several algorithmic optimizations such as weight clipping and scaling factor search, leading to SOTA model performance under 4-bit and 3-bit weight precision. We evaluate the model performance when applying AWQ and OmniQuant optimizations on top of the \textit{BitMoD} data type. We also compare with SOTA software-only LLM quantization methods, including GPTQ~\cite{gptq} and QuaRot~\cite{quarot} under weight-only quantization. Table~\ref{tab:sota_quant} shows the Wikitext and C4 perplexity of the studied Llama models using different quantization strategies. Combining \textit{BitMoD} with AWQ and OmniQuant significantly outperforms other approaches. For example, applying the \textit{BitMoD} data type reduces the average perplexity loss of OmniQuant by $28\%$ and $31\%$ at 4-bit and 3-bit precision, respectively. Overall, \textit{BitMoD} combined with AWQ and OmniQuant achieves an average perplexity loss of ${<1}$ for both 4-bit and 3-bit weight precision, pushing the limit of LLM weight quantization to a new state-of-the-art. It's important to note that using AWQ and OmniQuant does not inhibit the functionality of the \textit{BitMoD} accelerator---their optimization merely adjusts the per-group scaling factor, which is supported by the bit-serial dequantization unit of \textit{BitMoD}.

\vspace{4pt}
\noindent
\textbf{Orthogonal to Activation Quantization. } 
SmoothQuant can quantize LLM activation to \texttt{INT8} with low accuracy loss. We conduct weight PTQ using \textit{BitMoD} and \texttt{INT-Asym} data types on the pre-calibrated Llama models from SmoothQuant that use \texttt{INT8} activation. Table~\ref{tab:smoothquant} shows the WikiText-2 perplexity under \texttt{FP16} and \texttt{INT8} activation using different quantized weight precision and data types. The perplexity improvement of \textit{BitMoD} over \texttt{INT-Asym} remains after quantizing activation to \texttt{INT8} using SmoothQuant, and the improvement is particularly pronounced at lower precision (i.e., 3-bit). For instance, on Llama-3-8B, \textit{BitMoD} improves the perplexity by $15.21$ compared to \texttt{INT3-Asym} after applying SmoothQuant. Notably, on \text{Llama-2-7B}, the perplexity of \textit{BitMoD} weight with \texttt{INT8} activation is even better than \texttt{INT-Asym} weight with \texttt{FP16} activation, which demonstrates the potential of \textit{BitMoD} for further reducing the LLM memory footprint under a target model quality.

    \begin{table} [t]
      \centering
      \setlength{\tabcolsep}{3.5pt}
      \renewcommand{\arraystretch}{1.15}
      \footnotesize
      \caption{Wikitext-2 perplexity ($\downarrow$) when activation maintains in FP16 or is quantized to INT8 with SmoothQuant (SQ8). For weights, we use per-group quantization with a group size of 128.}
      \vspace{-3pt}
        \begin{tabular}{ c c  c c c  c c c  c c }
            \Xhline{0.3ex}
              \textbf{Weight} & \textbf{Weight} & \multicolumn{2}{c}{\textbf{Llama-2-7B}} & \multicolumn{2}{c}{\textbf{Llama-2-13B}} & \multicolumn{2}{c}{\textbf{Llama-3-8B}}
              \\
              \textbf{Precision} & \textbf{Datatype} & FP16 & SQ8 & \;FP16 & SQ8\; & FP16 & SQ8  
              \\
            \Xhline{0.3ex}
              \multirow{1}{*}{8-bit} & INT8 & 5.47 & 5.52 & \;4.95 & 4.93\; & 6.13 & 6.26 
              \\
            \hline
              \multirow{2}{*}{4-bit} & INT4-Asym & 5.77 & 5.83 & \;5.01 & 5.09\; & 6.84 & 7.05 
              \\
              & \textit{BitMoD} & 5.72 & 5.76 & \;5.01 & 5.07\; & 6.73 & 6.87 
              \\
            \hline
              \multirow{2}{*}{3-bit} & INT3-Asym & 7.08 & 7.58 & \;5.64 & 5.99\; & 13.26 & 25.78
              \\
              & \textit{BitMoD} & 6.55 & 6.85 & \;5.5 & 5.82\; & 9.09 & 10.57 
              \\
            \Xhline{0.3ex}
        \end{tabular}
      \label{tab:smoothquant}
      \vspace{-5pt}
    \end{table}


\section{Related Work}
\noindent
\textbf{DNN Accelerators. } 
There is an abundance of prior work on DNN accelerators~\cite{stripes, pragmatic, laconic, fpraker, bitwave, bbs, olaccel, gobo, mokey, ant, olive, cambricon-s, sparten, dstc, eureka, highlight, microscaling, figna}. These accelerators propose specialized processing elements and data flow to match the computational characteristics and memory access pattern of DNNs. Some accelerators exploit value sparsity to accelerate small-scale DNNs with the help of retraining~\cite{cambricon-s, sparten, dstc, eureka, highlight}. Other works target low-precision DNN acceleration based on model quantization~\cite{olaccel, gobo, mokey, ant, olive, microscaling, figna}. Among them, \cite{ant, olive, microscaling} introduce custom data types to better fit the value distribution of DNNs. Another line of works relies on bit-serial computing to scale the performance with lower operand precision, and leverages bit-level sparsity to skip ineffectual bit operations~\cite{stripes, pragmatic, laconic, fpraker, bitwave, bbs}. The proposed \textit{BitMoD} combines the benefits of quantization and bit-serial computing to efficiently trade-off between weight precision and hardware efficiency.

\vspace{4pt}
\noindent
\textbf{LLM Quantization. } Numerous algorithmic studies have proposed quantization solutions to reduce the memory footprint of LLMs~\cite{gptq, awq, llm-int8, students, llm-qat, zeroq_llm, smoothquant, quarot, omniquant, efficientqat}. Most of these works rely on asymmetric integer quantization for LLM weights while applying other techniques to optimize the quantization parameters. The proposed \textit{BitMoD} data type is orthogonal to many of these works and can be synergistically combined with different quantization optimizations.

\section{Conclusion}
In this paper, we introduce \textit{BitMoD}, an algorithm-hardware co-design scheme for efficient LLM acceleration. On the algorithm side, \textit{BitMoD} designs new data types that are tailored for per-group LLM weight quantization at very low precision. 
By intelligently repurposing the redundant zero value to an additional number, \textit{BitMoD} extends the resolution or range of 3-bit-and 4-bit floating-point data types. 
Moreover, the \textit{BitMoD} quantization framework can be seamlessly integrated with existing software-only quantization methods to further improve the model performance. On the hardware side, \textit{BitMoD} proposes a unified bit-serial representation for diverse low-precision data types and an efficient bit-serial PE to process quantized weight and FP16 activation. Our evaluation demonstrates that \textit{BitMoD} significantly outperforms existing LLM quantization methods, pushing the limit of LLM weight quantization to a new state-of-the-art. Compared to prior accelerators ANT\,/\,OliVe, \textit{BitMoD} achieves $1.69\times$\;/\;$1.48\times$ speedup and $1.48\times$\;/\;$1.31\times$ better energy efficiency, while being able to support diverse weight precision to offer a good trade-off between model accuracy and hardware efficiency.

\section*{Acknowledgment}
This project is supported in part by Intel Corporation, the National Science Foundation under Grant No. 2339084, and the Engineering and Physical Sciences Research Council (EPSRC) under Grant No. EP/S030069/1.
We would like to thank Jordan Dotzel, Hongxiang Fan, Stylianos I. Venieris, Alexandros Kouris, Mahesh Iyer, Grace Zgheib, Sergey Gribok, and the anonymous reviewers for their constructive feedback. 


\bibliographystyle{IEEEtranS}
\bibliography{refs}

\clearpage
\appendix
\section{Artifact Appendix}

\subsection{Abstract}
This artifact appendix describes how to access and evaluate the BitMoD artifact~\cite{bitmod-ae} to reproduce the experiments as performed in Section~\ref{sec:evaluation}. 

\subsection{Artifact Check-List (Meta-Information)}
\begin{itemize}
  \item {\bf Program:} Python, Shell script
  \item {\bf Runtime Environment:} Ubuntu 20.04
  \item {\bf Hardware:} NVIDIA GPU with $\geq$40 GB of VRAM (e.g., A6000).
  \item {\bf Model (LLM):} OPT-1.3B, Phi-2B, Yi-6B, Llama-2-7B, Llama-2-13B, Llama-3-8B.
  \item {\bf Dataset:} Wikitext, C4.
  \item {\bf Experiments:} LLM perplexity evaluation in Section~\ref{sec:accuracy} and Section~\ref{sec:comb_sota_quant}. Accelerator performance evaluation in Section~\ref{sec:hardware_eval}.
  \item {\bf Disk space required (approximately):} 512 GB.
  \item {\bf How much time is needed to complete experiments (approximately)?:} 50 hours.
  \item {\bf Publicly available?:} Yes.
  \item {\bf Archived:} https://doi.org/10.5281/zenodo.14252531
\end{itemize}

\subsection{Description}

\subsubsection{How to Access}
We maintain a publicly-available repository~\cite{bitmod-ae} where we have open-sourced all of our artifacts.

\subsubsection{Hardware Dependencies}
One NVIDIA GPU with at least 40 GB of VRAM (e.g., A6000), in addition to a normal desktop computer with at least 512 GB of free disk space.

\subsubsection{Software Dependencies}
All experiments require Conda for managing virtual Python environments. The quantization experiments require CUDA. Other requirements are automatically installed by scripts in the following sections. When running experiments, please use a tmux session to make sure long-running jobs are not killed unexpectedly. 
\subsection{Installation}
For artifact evaluation, begin by downloading the top-level repository from Zenodo:
\begin{lstlisting}[language=bash]
    $ wget -O BitMoD.zip https://zenodo.org/records/14252531/files/BitMoD-HPCA-25.zip
    $ unzip BitMoD.zip
\end{lstlisting}

The artifact is inside the unzipped \textbf{BitMoD-HPCA-25} repository, which contains five sub-folders, each targeting one set of experiments: 

\begin{enumerate}
    \item \textbf{bitmod-quant}, which runs the baseline weight-only quantization with different data types. This can reproduce the results in Table~\ref{tab:gen_task_ppl} and Table~\ref{tab:dtype_ablation}.
    \item \textbf{bitmod-sim}, contains a custom simulator that calculates the latency and energy of the \textit{BitMoD} accelerator. This can reproduce the results in Fig.~\ref{fig:speedup} and Fig.~\ref{fig:energy}.
    \item \textbf{AWQ-BitMoD}, which runs AWQ~\cite{awq} with integer and \textit{BitMoD} data types. This can reproduce the AWQ results in Table~\ref{tab:sota_quant}.
    \item \textbf{OmniQuant-BitMoD}, which runs OmniQuant~\cite{omniquant} with integer and \textit{BitMoD} data types. This can reproduce the OmniQuant results in Table~\ref{tab:sota_quant}.
    \item \textbf{SmoothQuant-BitMoD}, which runs SmoothQuant~\cite{smoothquant} with integer and \textit{BitMoD} data types for weight quantization. This can reproduce the results in Table~\ref{tab:smoothquant}.
\end{enumerate}

Please go to every sub-folder and refer to the corresponding `\texttt{README.md}' for detailed setup instructions. Note that AWQ, OmniQuant, and SmoothQuant will require different conda environments. Hence, please change back to the base environment after installing a conda environment before creating the next.


For example, the AWQ environment can be created by running:
\begin{lstlisting}[language=bash]
    $ cd AWQ-BitMoD
    $ conda create -n awq-bitmod python=3.10 -y
    $ conda activate awq-bitmod
    # Follow `README.md' inside AWQ-BitMoD folder to set up other dependencies.
    $ conda deactivate  # change back to the base environment
\end{lstlisting}

The OmniQuant and SmoothQuant environments can be created in a similar way by repeating the above step and following their `\texttt{README.md}' to set up other dependencies.

\subsection{Experiment workflow}
Once the environment is set up, we will conduct five sets of experiments, each corresponding to one of the five sub-folders within the \textbf{BitMoD-HPCA-25} repository.

\vspace{6pt}
\subsubsection{BitMoD Weight-only Quantization} \label{sec:ae_weight_only}
Run the basic LLM weight-only quantization experiments to reproduce the results in Table~\ref{tab:gen_task_ppl} and Table~\ref{tab:dtype_ablation}.
\begin{lstlisting}[language=bash]
    $ cd bitmod_quant
    $ conda activate awq-bitmod 
\end{lstlisting}

In `\texttt{run\_exp.sh}', modify the `\texttt{export}' command by specifying the HuggingFace home directory, `\texttt{HF\_HOME}', on your computer. By default, this can be set to your home directory. 
\begin{lstlisting}[language=bash]
    $ export HF_HOME="your/HF_HOME/directory"
\end{lstlisting}

Then run the following: 
\begin{lstlisting}[language=bash]
    $ bash run_exp.sh
\end{lstlisting}
The perplexity result will be saved in the folder called `\texttt{results\_quant}'.

\vspace{6pt}
\subsubsection{BitMoD Hardware Simulation}
Before running the simulator, go to `\texttt{bitmod\_sim}' of the repository:
\begin{lstlisting}[language=bash]
    $ cd bitmod_sim
    $ conda activate awq-bitmod 
\end{lstlisting}

In `\texttt{run\_shape\_profile.sh}', modify the `\texttt{export}' command by specifying the HuggingFace home directory, `\texttt{HF\_HOME}', on your computer:
\begin{lstlisting}[language=bash]
    $ export HF_HOME="your/HF_HOME/directory"
\end{lstlisting}

Then generate the model shape information that can be passed to the accelerator simulator: 
\begin{lstlisting}[language=bash]
    $ bash run_shape_profile.sh
\end{lstlisting}

Next, run different simulators for the baseline FP16 accelerator, ANT, OliVe, and \textit{BitMoD}:
\begin{lstlisting}[language=bash]
    $ python test_baseline.py --is_generation 
    $ python test_ant.py      --is_generation 
    $ python test_olive.py    --is_generation  
    $ python test_bitmod.py   --is_generation --is_lossless 
\end{lstlisting}
The flag \textit{-{}-is\_generation} is optional. When enabled / disabled, it will evaluate the hardware performance of generative / discriminative tasks. The flag \textit{-{}-is\_lossless} is optional for \textit{BitMoD}. When enabled / disabled, it will evaluate the hardware performance of lossless / lossy \textit{BitMoD} quantization. 

Finally, to generate  Fig.~\ref{fig:speedup} and Fig.~\ref{fig:energy} of the paper, go to `\texttt{bitmod\_sim/plot}' directory and run the Jupyter notebooks inside. Note that the cycle and energy numbers are the same as those output by the simulators.

\vspace{6pt}
\subsubsection{AWQ}
Go to the `\texttt{AWQ-BitMoD}' directory:
\begin{lstlisting}[language=bash]
    $ cd AWQ-BitMoD
    $ conda activate awq-bitmod 
\end{lstlisting}

In `\texttt{run\_awq.sh}' and `\texttt{run\_eval\_ppl.sh}', modify the first `\texttt{export}' command by specifying the HuggingFace home directory, `\texttt{HF\_HOME}', on your computer:
\begin{lstlisting}[language=bash]
    $ export HF_HOME="your/HF_HOME/directory"
\end{lstlisting}

Then, run the following two commands separately:
\begin{lstlisting}[language=bash]
    $ bash run_awq.sh  # will take several hours
    $ bash run_eval_ppl.sh 
\end{lstlisting}
The perplexity results will be saved in the folder called `\texttt{results}'. You can compare these with the AWQ results in Table~\ref{tab:sota_quant}.

\vspace{6pt}
\subsubsection{OmniQuant}
Go to the `\texttt{OmniQuant-BitMoD}' directory:
\begin{lstlisting}[language=bash]
    $ cd OmniQuant-BitMoD
    $ conda activate omniquant-bitmod 
\end{lstlisting}

The comprehensive scripts to reproduce the Table~\ref{tab:sota_quant} OmniQuant results are available in the `\texttt{scripts}' directory. Before running any command in the scripts, execute the following `\texttt{export}' command and specify the HuggingFace home directory, `\texttt{HF\_HOME}', on your computer:
\begin{lstlisting}[language=bash]
    $ export HF_HOME="your/HF_HOME/directory"
\end{lstlisting}

In every shell script, you need to change the parameter of \textit{-{}-model} flag to the LLM path in your computer. By default, this can be set to the LLM directory from the official HuggingFace website. For example, inside the script `\texttt{llama-2-13b-int.sh}', the Llama-2-7B model path can be specified with:
\begin{lstlisting}[language=bash]
    --model meta-llama/Llama-2-7b-hf 
\end{lstlisting}

After changing the \textit{-{}-model} flag to the correct model path, copy and execute every script's python command under the `\texttt{OmniQuant-BitMoD}' directory. You may check the perplexity results at the end of the log file specified by the \text{\textit{-{}-output\_dir}} flag in every command, and compare those with the OmniQuant results in Table~\ref{tab:sota_quant}.

\vspace{6pt}
\subsubsection{SmoothQuant}
Go to the `\texttt{SmoothQuant-BitMoD}' directory:
\begin{lstlisting}[language=bash]
    $ cd SmoothQuant-BitMoD
    $ conda activate smoothquant-bitmod 
\end{lstlisting}

In `\texttt{run\_experiments.sh}', modify the `\texttt{export}' command by specifying the HuggingFace home directory, `\texttt{HF\_HOME}', on your computer:
\begin{lstlisting}[language=bash]
    $ export HF_HOME="your/HF_HOME/directory"
\end{lstlisting}

Then, run the following command:
\begin{lstlisting}[language=bash]
    $ bash run_experiments.sh
\end{lstlisting}
The perplexity results will be saved in the folder called `\texttt{results\_mod}'. You can compare these results with the SmoothQuant results in Table~\ref{tab:smoothquant}.

\subsection{Evaluation and expected results}
There are five result folders after running the above experiments:
\begin{enumerate}
    \item \texttt{bitmod-quant/results\_quant}, contains the perplexity results in Table~\ref{tab:gen_task_ppl} and Table~\ref{tab:dtype_ablation}.
    \item \texttt{bitmod-sim/plot}, contains two Jupyter notebooks to reproduce Fig.~\ref{fig:speedup} and Fig.~\ref{fig:energy}, respectively.
    \item \texttt{AWQ-BitMoD/results}, contains the AWQ results in Table~\ref{tab:sota_quant}.
    \item \texttt{OmniQuant-BitMoD/log}, contains the OmniQuant results in Table~\ref{tab:sota_quant}.
    \item \texttt{SmoothQuant-BitMoD/results\_mod}, contains the SmoothQuant results in Table~\ref{tab:smoothquant}.
\end{enumerate}

\subsection{Methodology}
Submission, reviewing and badging methodology:

\begin{itemize}[leftmargin=7ex]
  \item \url{https://www.acm.org/publications/policies/artifact-review-and-badging-current}
  \item \url{https://cTuning.org/ae}
\end{itemize}


\end{document}